\documentclass{article}
\usepackage[utf8]{inputenc}
\usepackage{amsthm}
\usepackage{amsmath}
\usepackage{amssymb}
\usepackage[a4paper, total={6in, 9in}]{geometry}
\usepackage[linesnumbered,ruled,vlined]{algorithm2e} 
\theoremstyle{definition}

 \usepackage{graphicx}
\theoremstyle{remark}
\newtheorem*{remark}{Remark}
\usepackage[T1]{fontenc}

\title{Where is the Bottleneck of Adversarial Learning with Unlabeled Data?}
\author{Jingfeng Zhang$^{*}$ \\  {j-zhang@comp.nus.edu.sg} 
   \and Bo Han\footnote{Equal contributions.} \\  {bo.han@riken.jp} \and
   Gang Niu \\ {gang.niu@riken.jp} \and 
   Tongliang Liu \\{tongliang.liu@sydney.edu.au} \and 
   Masashi Sugiyama \\{sugi@k.u-tokyo.ac.jp}}

\date{}

\usepackage{graphicx}

\begin{document}

\maketitle
\begin{abstract}
Deep neural networks (DNNs) are incredibly brittle due to adversarial examples. To robustify DNNs, adversarial training was proposed, which requires large-scale but well-labeled data. However, it is quite expensive to annotate large-scale data well. To compensate for this shortage, several seminal works are utilizing large-scale unlabeled data. In this paper, we observe that seminal works do not perform well, since the quality of pseudo labels on unlabeled data is quite poor, especially when the amount of unlabeled data is significantly larger than that of labeled data. We believe that the quality of pseudo labels is the bottleneck of adversarial learning with unlabeled data. To tackle this bottleneck, we leverage deep co-training, which trains two deep networks and encourages two networks diverged by exploiting peer's adversarial examples. Based on deep co-training, we propose robust co-training (RCT) for adversarial learning with unlabeled data. We conduct comprehensive experiments on  \textit{CIFAR-10} and \textit{SVHN} datasets. Empirical results demonstrate that our RCT can significantly outperform baselines (e.g., robust self-training (RST)) in both standard test accuracy and robust test accuracy w.r.t. different datasets, different network structures, and different types of adversarial training.
\end{abstract}

\section{Introduction}
Due to their superior performance, deep neural networks (DNNs) have been deployed on real systems in many fields, such as image recognition~\cite{he2016deep} and natural language processing~\cite{zhang2015character}. Real-world systems could take inputs shifted by various perturbations, e.g., different lighting effects on an image, various ambient noise on a conversation. Those could potentially cause unreliable predictions of DNNs. In particular, crafted adversarial examples~\cite{szegedy} can easily flip the predictions of deployed DNNs, through adding imperceptible noise to natural data. It arouses anxieties on deploying DNNs in safety-critical fields, such as autonomous driving~\cite{chen2015deepdriving} and medical images analysis~\cite{litjens2017survey}.

Recently, many efforts have been made on learning robust DNNs to resist such adversarial examples. In general, there are two broad branches in adversarial machine learning, i.e., certified robust training \cite{wong2017provable, tsuzuku2018, JeremyCohenCertifiedML, LecuyerAdv_DP} and empirical robust training \cite{MadryMSTV18,Zhang_trades,Maxingjun2019convergence}. Their common purpose is to construct robust DNNs to mimic the natural occurring system (e.g., human visual system). A system is believed to be robust and invariant to adversarial perturbations since its output is smooth w.r.t. its input~\cite{wahba1990spline}.

To acquire such smoothness, we can conduct data augmentation using adversarial data~\cite{MadryMSTV18,Maxingjun2019convergence} or perturbed data with Gaussian random noise~\cite{litjens2017survey,JeremyCohenCertifiedML} during training. In this way, predictions of DNNs around the data input could be insensitive to imperceptible perturbations. 
Nonetheless, Dimitris et al.~elucidated that adversarial robustness may be at odds with standard accuracy~\cite{Tsipras_adversarial_at_odd}. To mitigate the large gap between robustness and accuracy, more well-labeled samples are needed during training~\cite{schmidt2018adversarially}, which also achieves the greater smoothness close to that of a natural occurring system. However, it is expensive to gather well-labeled data, not to mention large-scale well-labeled data for the smoothness requirement~\cite{schmidt2018adversarially}. Fortunately, this issue can be alleviated through seminal efforts~\cite{Miyato,Percy,DeepMind}, namely utilizing unlabeled data to improve adversarial robustness. Conceptually, above works consist of three components: 
\begin{itemize}
    \item[] (a) Given the existing training data $S_L = \{ (\mathbf{x}_i, y_i ) \}^{N_L-1}_{i=0}$ where $(\mathbf{x}_i, y_i ) \sim P (X,Y) $, they collected extra unlabeled data $D_U = \{\mathbf{x}_j \}^{N_U-1}_{j = 0 }$ where $\mathbf{x}_j \sim P(X)$. For example, to obtain the distribution of unlabeled data similar to that of \textit{CIFAR-10}~\cite{krizhevsky2009learning_cifar10}, \textit{80 million tiny images}~\cite{torralba200880_tiny_image} could be utilized, where \textit{CIFAR-10} is a subset annotated by human. 
    \item[] (b) Based on those existing training data $S_L$, they annotated unlabeled data $D_U$ to get $S_U = \{(\mathbf{x}_j, \overline{y}_j)\}^{N_U-1}_{j=0}$, where $(\mathbf{x}_j, \overline{y}_j) \sim \overline{P}(X, \overline{Y})$, $\overline{P}(X) = P(X)$ and $\overline{P}(\overline{Y}|X) \neq P(Y|X)$. The goal is to minimize the divergence (e.g., KL divergence) between $\overline{P}(\overline{Y}|X)$ and $P(Y|X)$.
    \item[] (c) By jointly using dataset $S = S_L \cup S_U$, they train robust DNNs using existing strategies, such as Madry's adversarial training~\cite{MadryMSTV18}, adversarial training TRADES~\cite{Zhang_trades}, and random smoothing~\cite{JeremyCohenCertifiedML}.
\end{itemize}

For example, UAT++ in~\cite{DeepMind} and SSDRL in~\cite{Miyato} integrate parts (b) and (c) into the objective functions of DNNs, which encourages the model output on unlabeled data close to unknown ground-truth labels when number of labeled training samples is large. However, the integration limits the diversity of annotation methods (i.e., part (b)), and it only enables regularization-based methods (e.g.,VAT~\cite{MiyatoVAT}) to annotate extra unlabeled data. Many potential methods are excluded, such as co-training~\cite{blum1998combining,qiao2018deepcotraining} and graph-based models~\cite{zhou2005Graphbasedlearning}. By contrast, robust self-training (RST) in~\cite{Percy} has three independent modules for parts (a)--(c), and each fungible part has its clear purpose. Thus, it is believed to be the best by the standard of modular design~\cite{westra2016modular}.

The purpose of part (a) is to gather qualified unlabeled data as much as possible, e.g., scratch websites for unlabeled images or collect medical images without doctor's diagnosis. To gather such data, there are many standard methods, thus the improvement of this part is out the scope of the current paper. Meanwhile, different training methods in part (c) seem to hit their limits, which hardly narrows the gap between robust generalization and standard generalization further~\cite{schmidt2018adversarially}. For example, on CIFAR-10, the state-of-the-art TRADES achieves the robust accuracy around 50\% \cite{Zhang_trades}, while the standard accuracy should be above 90\%~\cite{he2016deep,Jingfeng_robust_resnet}.

\begin{figure}[h!]
	\centering
	\includegraphics[width=0.31\linewidth] {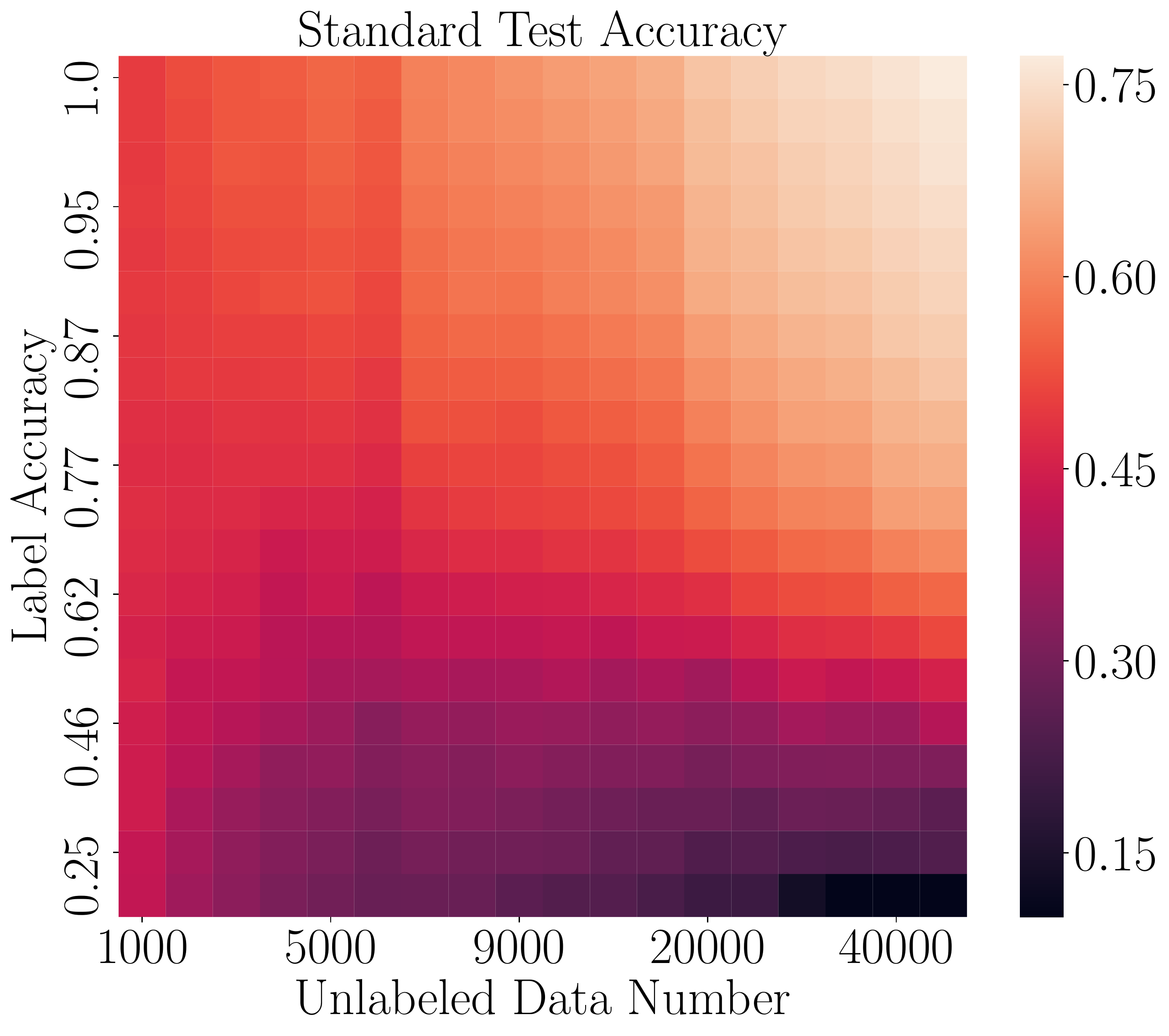}
	\includegraphics[width=0.31\linewidth]{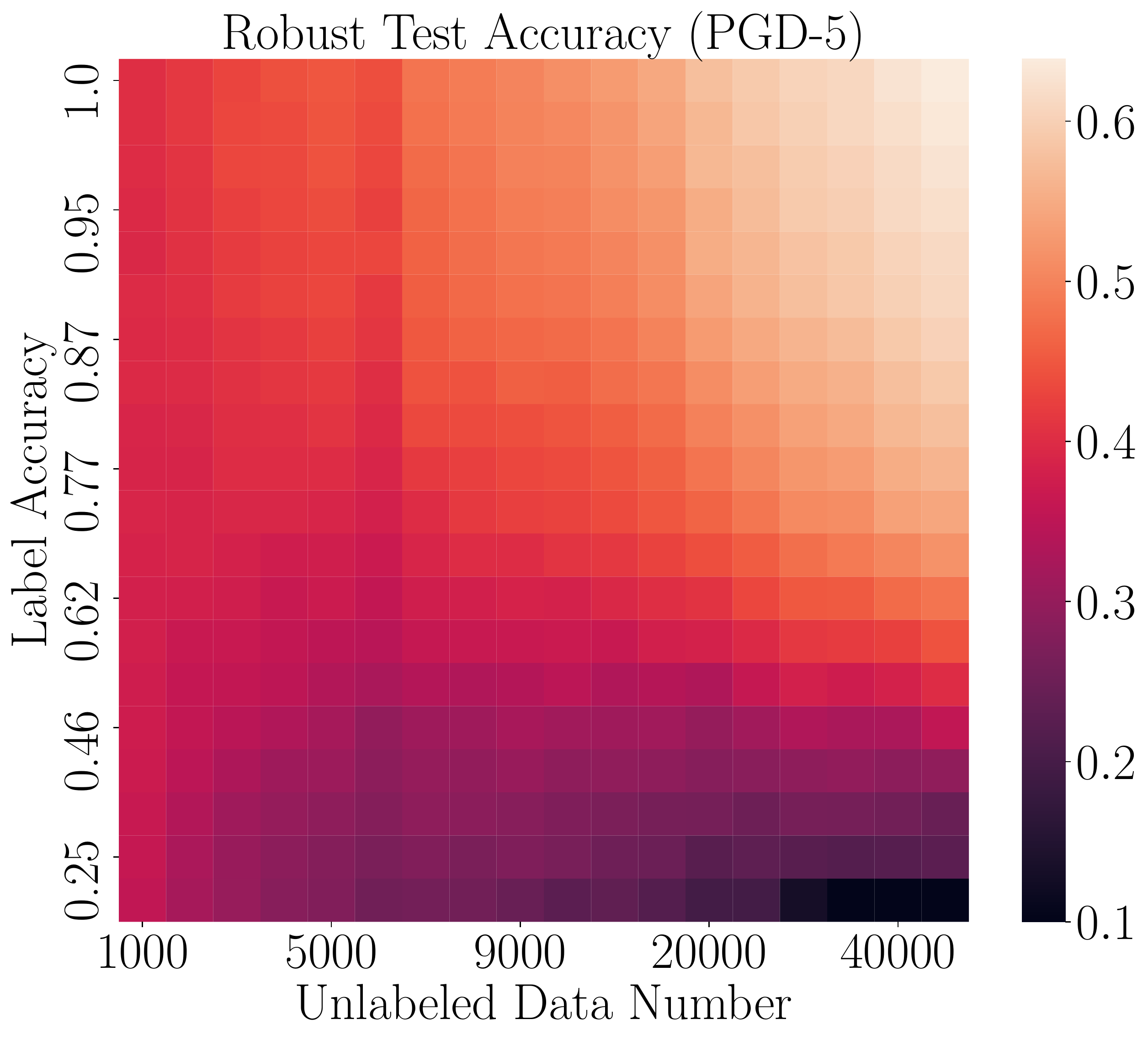}
	\includegraphics[width=0.31\linewidth]{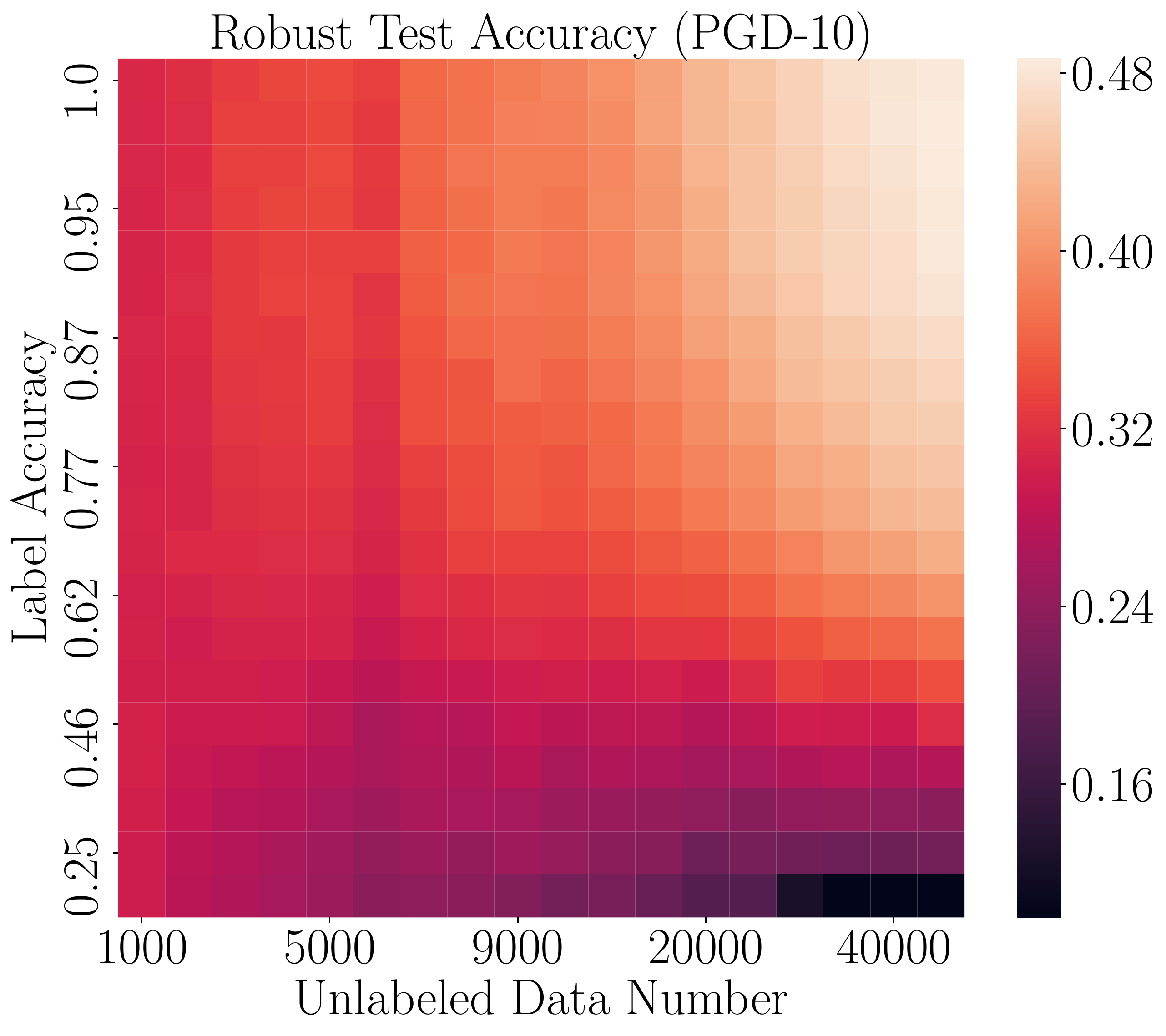} \\
	\includegraphics[width=0.31\linewidth] {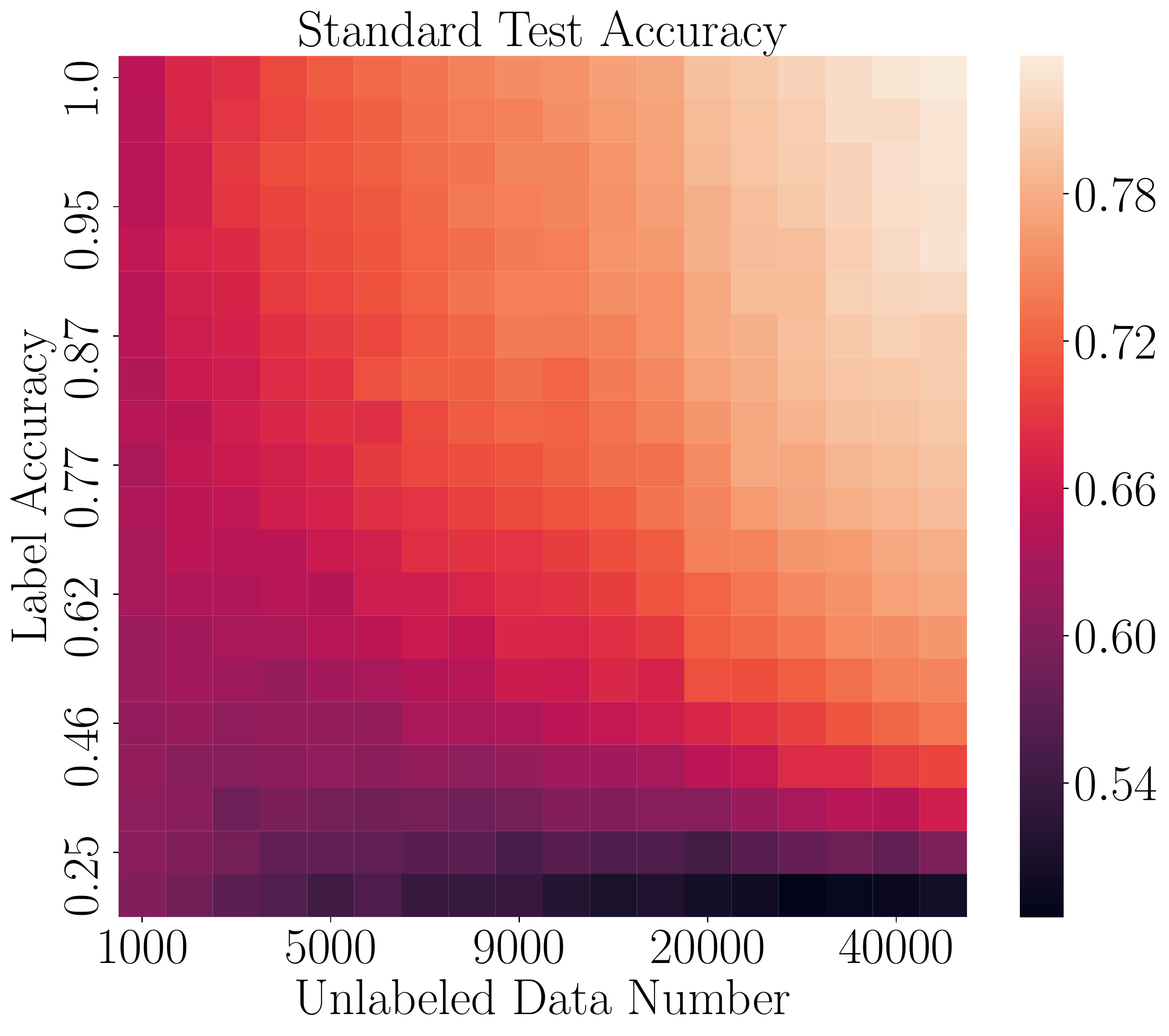}
	\includegraphics[width=0.31\linewidth]{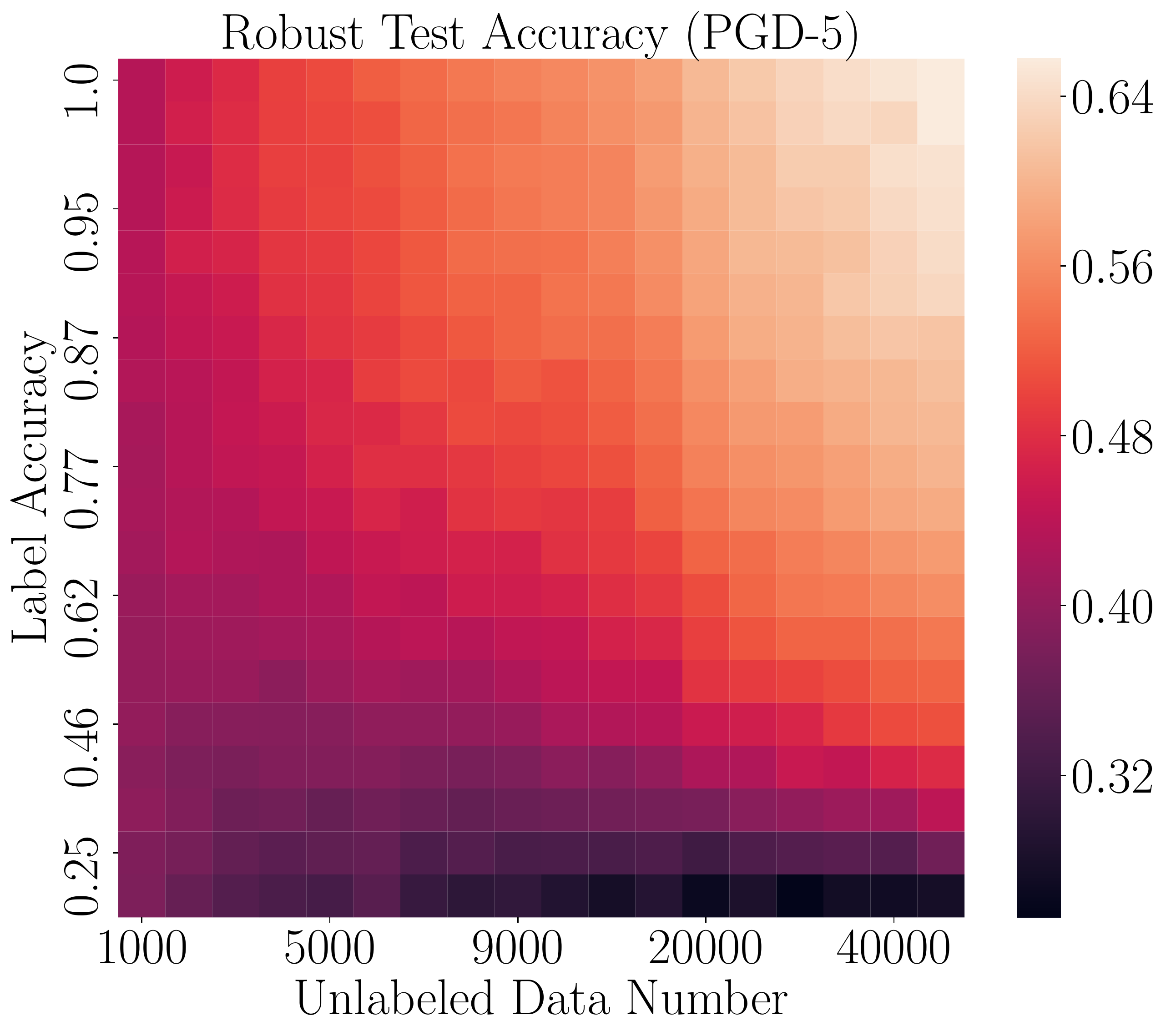}
	\includegraphics[width=0.31\linewidth]{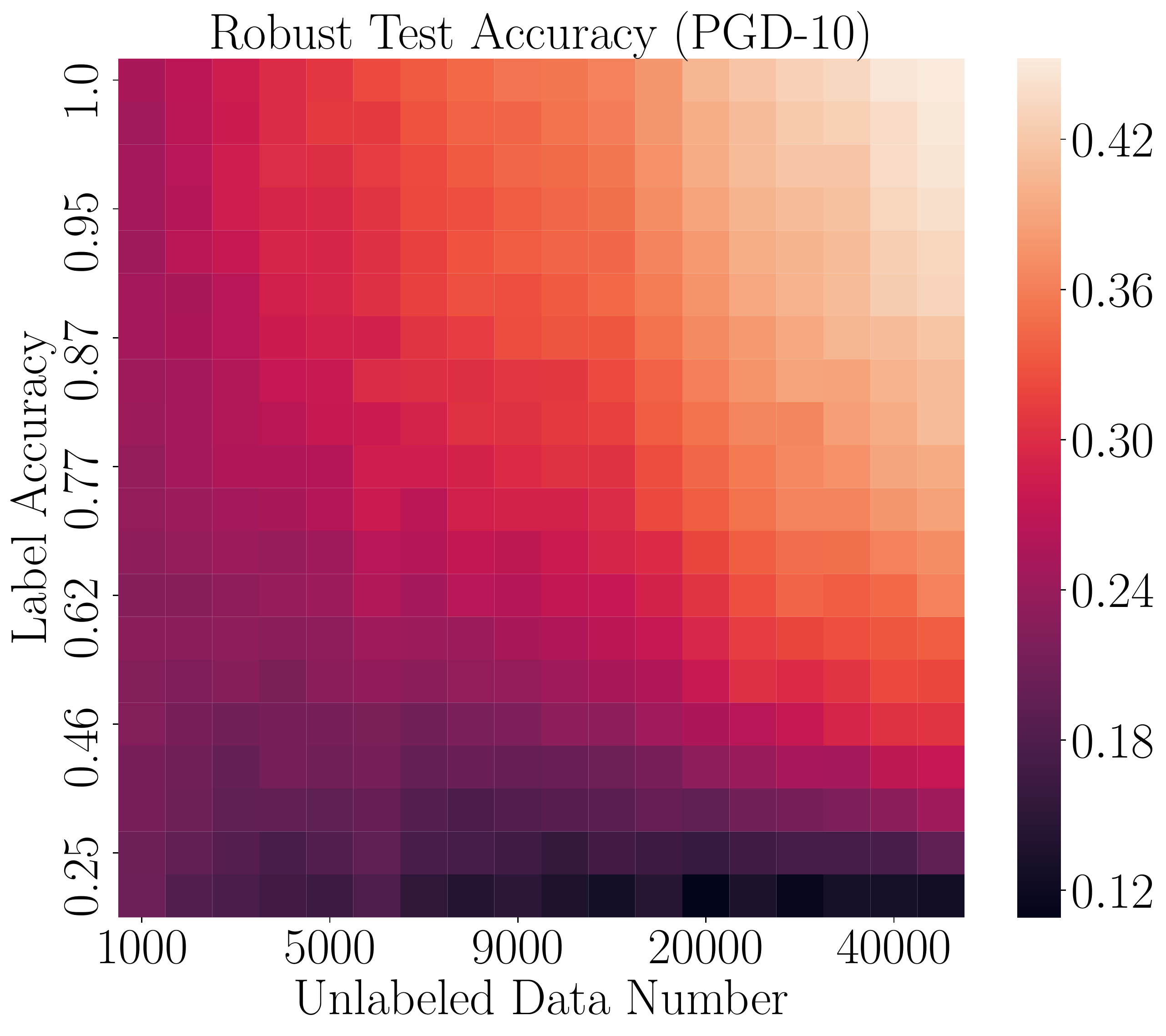}
	\caption{With $|S_L| = 4k$ out of $50k$ \textit{CIFAR-10} training data separated out as labeled dataset and the remaining $46k$ treated as unlabeled dataset $D_U$, we compare the adversarial training performance, i.e., standard test accuracy and robust test accuracy. In each cell we conduct an adversarial training. For example, for the cell with unlabeled data number 20000 and label accuracy 0.87, we adversarially train a ResNet10~\cite{he2016deep} based on training dataset $S = S_L \cup S_U$, where $S_U$ is pseudo label dataset of $D_U$ and its labels' accuracy is 87\%. After adversarial training, we conduct the evaluation using $10k$ \textit{CIFAR-10} test data. Standard test accuracy is evaluated on the natural data. Robust test accuracy (PGD-5) and robust test accuracy (PGD-10) are evaluated on adversarial data generated from its corresponding natural data using PGD-5 and PGD-10~\cite{MadryMSTV18} respectively. We use ResNet10 for all adversarial training. 
	The three panels above use Madry's adversarial training~\cite{MadryMSTV18}. The three panels below use adversarial training TRADES~\cite{Zhang_trades}. }
	\label{label_purity_heat_map}
\end{figure}

Thus, this motivates us to improve part (b), namely including more well-labeled data. The label quality is quite crucial to boosting the adversarial robustness. For example, in Figure~\ref{label_purity_heat_map}, given the fixed amount of unlabeled data, the increased label accuracy boosts both standard accuracy and robust accuracy significantly. As the amount of unlabeled data increases, high label accuracy takes positive effects on adversarial robustness, vice versa. Meanwhile, negative effects of low-quality labels are reinforced during training. RST~\cite{Percy} firstly learns a classifier based merely on labeled dataset $S_L$, then uses the learned classifier to annotate all unlabeled data $D_U$ with pseudo labels to get $S_U$. We name this classifier pre-determined annotator. The pre-determined annotator does not consider the knowledge of unlabeled data. As an simple example in Figure~\ref{SSL_benefits} illustrates, an annotator based solely on the labeled data $S_L$ may give wrong labels to a large portion of unlabeled data. RST has a bottleneck that it could give poor pseudo labels to unlabeled data, and later adversarial training could be fed on many erroneous data. Even worse, its error is accumulated and reinforced over training. The quality of pseudo labels decides the success of adversarial training. 

Fortunately, there remains a lot of room to improve the quality of these labels. To break the bottleneck of RST~\cite{Percy}, we leverage deep co-training to improve the quality of pseudo labels in part (b), and thus propose robust co-training (RCT) for adversarial learning with unlabeled data. The proposed algorithm utilizes two networks to correct the mistake of each other by getting consensus on unlabeled data. Meanwhile, each network robustly trains on adversarial examples generated by its peer network, which keeps both networks diverged in function. Our experiments confirm its effectiveness on the quality of pseudo labels, which could further boost both standard test accuracy and robust test accuracy in adversarial training. Our proposed method takes a giant leap in closing the gap between adversarially robust generalization and standard generalization.

\section{Related Work}

\subsection{Semi-supervised Deep Learning}
% \footnote{+++ please reorganize your work by 
% 1) https://arxiv.org/pdf/1804.09170.pdf; 2) https://arxiv.org/pdf/1905.02249.pdf}

Many works have been proposed to boost the label quality of unlabeled data largely located in the area of semi-supervised learning (SSL). Self-training~\cite{mcclosky2006effective_self_train,lee2013pseudo_self_train} is one of simplest approaches in SSL. Self-training produces pseudo labels for unlabeled data using the model itself to obtain additional training data. Unlabeled data with confident predictions are recruited into training. However, self-training is hardly able to correct its own mistakes. If the model's prediction on unlabeled data is confident but wrong, the wrong pseudo-labeled data is forever incorporated into training and it amplifies the model error over training iterations.

Multi-view training aims to train multiple models with different views of the data. These view enhance each other and can help to correct other's mistakes. 
The most exemplar one is co-training~\cite{blum1998combining}.
To be specific, in~\cite{blum1998combining} different views refer to different independent set of feature on the same data. For example, in web page classification, one set of feature is text on the webpage, another set of feature is its anchor text hyperlinks to that webpage. There are two models looking at different sets of feature.
Each model are trained on its respective feature set. Over training iterations, unlabeled data with confident predictions by one model are moved to training set of its peer model. 
%In~\cite{qiao2018deepcotraining}, different views of two models are encouraged by utilizing each other's virtual adversarial examples on the same data. 

Regularization based semi-supervised learning encourages output of different perturbations of input data to be close, through adding the regularization term in the loss function. For example, \cite{bachman2014learning_pi_model,Laine_TemporalEnsembling,tarvainen2017meanteacher} use random perturbations and \cite{MiyatoVAT} uses virtual adversarial perturbations. 
A comprehensive review on SSL, e.g. generated model based SSL and graph-based SSL refers to~\cite{Chapelle_SSL_BOOK}. 

\subsection{Adversarial Defense}
Many works focus on building adversarially robust models against adversarial perturbations. In general, those are divided into two branches certified defenses and empirical defenses.

In certified defenses, the model's prediction is expected to be unchanged for any perturbed data around its corresponding natural data.
% \footnote{+++ Sorry, I don't know what do you want to guarantee JF's ANS: revised.} 
There are some exemplar works \cite{Raghunathan_certified_defence,wong2017provable,JeremyCohenCertifiedML}.
For example, \cite{LecuyerAdv_DP,JeremyCohenCertifiedML} use randomized smoothing to  
% \footnote{+++ Could you please use the basic word? JF's ANS: Revised.}
transform base classifier to a new smoothed classifier.
However, due to its strong assumption, certified robustness has difficulty in scalability in large models and high dimensional data, and suffers from low computational efficiency in its robustness certification.
%random smoothing need to know base model's predictions on all surroundings of input. It is very difficult especially input dimension is high. 
%Due to its strong assumption, there is a limitation on DNN scalability and large dataset \footnote{+++ I don't think scalability and large dataset are parallel}. 

Another line of defense
% \footnote{+++ Could you please be careful about defense instead of defence? JF's ANS: Corrected.} 
is empirical defense.  Empirical defense dynamically exploits adversarial examples and recruit them into the training along with natural data. Adversarial examples are exploited according to natural data.  The network has a large loss on them, but they are visually indistinguishable with their natural data counterpart.
The most exemplar ones are Madry's adversarial training~\cite{MadryMSTV18} and adversarial training TRADES~\cite{Zhang_trades}.

In empirical defense, the purpose of defense is to minimize the adversarial risk, i.e., 
\begin{align*}
    R_{adv}(f_{\theta}) = \mathbf{E}_{(\mathbf{x},y) \sim \mathcal{D}} \displaystyle [\underset{\delta \in \triangle(\mathbf{x})}{\max} \mathcal{L} (f_{\theta} (\mathbf{x}+\delta), y)\displaystyle ]
\end{align*}
where $\mathcal{D}$ denotes the true distributions over samples and $\triangle(\mathbf{x})$ denoted the allowed perturbations region of the sample point.
The empirical defense is to find parameter $\theta$ minimize the empirical risk
\begin{align*}
    \hat{R}_{adv}(f_{\theta}, S_L) = \frac{1}{|S_L|}\displaystyle \sum_{(\mathbf{x},y) \in S_L} \displaystyle [\underset{\delta \in \triangle(\mathbf{x})}{\max} \mathcal{L} (f_{\theta} (\mathbf{x}+\delta), y) \displaystyle ]
\end{align*}
where $S_L$ is a finite set of samples drawn i.i.d. from $\mathcal{D}$.
To solve this min-max problem, ~\cite{MadryMSTV18} applies Danskin's theorem~\cite{bertsekas1997nonlinear_danskin}. At each training iteration, Madry's adversarial training firstly exploits adversarial examples that maximize the loss and then update the classifier based on these adversarial examples, i.e., 
\begin{align} \label{madry_adversarial_train}
    \theta =\underset{\theta}{\arg\min}   \frac{1}{|S_L|} \displaystyle \sum_{x,y \in S_L } \displaystyle [ \underset{|| \mathbf{x}' - \mathbf{x}|| \leq \epsilon }{\max}
\mathcal{L} (  f_{\theta}(\mathbf{x}'), y)  \displaystyle ]
\end{align}
where $\mathbf{x}'$ is adversarial example of $\mathbf{x}$ within its allowed perturbation region $\triangle(\mathbf{x})$, i.e., $||\mathbf{x}' - \mathbf{x}|| \leq \epsilon$. The inner maximization is non-convex optimization problem with difficulty to get its exact solution. Projected gradient descent (PGD)~\cite{MadryMSTV18} is utilized to approximately search its local minima. $\mathcal{L}$ is the cross-entropy loss encouraging the predicted value of the adversarial example $\mathbf{x}'$ to be near the true label of its corresponding natural example $\mathbf{x}$. 

Another exemplar work is TRADES~\cite{Zhang_trades}. Similar to VAT~\cite{MiyatoVAT}, they introduce a regularization term on the loss function encouraging similarity between predictions of $f(\mathbf{x})$ and its adversarial example $f(\mathbf{x}')$, i.e., 
\begin{align} \label{trades}
   \displaystyle \theta &=
   \underset{\theta}{\arg\min} \frac{1}{|S_L|}  \sum_{x,y \in S_L } \displaystyle [ \mathcal{L}  (  f_{\theta}(\mathbf{x}), y) \nonumber \\ &+  \underset{|| \mathbf{x}' - \mathbf{x}|| \leq \epsilon }{\max}
\lambda \mathcal{KL} (f_{\theta}(\mathbf{x}),  f_{\theta}(\mathbf{x}')) \displaystyle ]
\end{align}

%\footnote{+++ The math in Eq. (2) is correct, and I double check already. However, could you please explain it intuitively?}
%\footnote{+++ Why Trades uses KL but our use JS?}

where $\mathcal{KL}$ is the Kullback Leibler divergence which measure the prediction difference, $\mathcal{L}$ is cross-entropy loss, and $\lambda$ is the trade off parameter. It also uses PGD to approximately solve the inner maximization.

\section{Methodology}

In order to achieve greater smoothness in adversarial training, three seminal works leverage large-scale unlabeled data \cite{Miyato,Percy,DeepMind}. Figure~\ref{label_purity_heat_map} shows that, given a fixed amount of unlabeled data, both standard test accuracy and robust test accuracy can get improved when its label accuracy of pseudo labels improves. Thus, it is inevitable to require high-quality pseudo labels on those unlabeled data, and part (b) plays an vital role in the success of adversarial training.

Although Carmon et al. leverage the term ``robust self-training'' (RST) to characterize their algorithm \cite{Percy}, their actual operation for part (b) is not the conventional self-training. Specifically, they train a classifier merely based on $S_L$. Then, they use the pre-trained classifier to annotate all unlabeled data $D_U$ in \textit{one time}, which acquires pseudo labels on unlabeled data to get $S_U$. We name such pre-trained classifier as pre-determined annotator. Finally, they jointly use dataset $S = S_L \cup S_U$ to train a adversarially robust deep neural network. However, is the pre-determined annotator good enough to annotate unlabeled dataset $D_U$? The answer is negative.

\begin{figure}[!tp]
 \vspace{-5mm}
	\centering
	\includegraphics[scale=0.35]{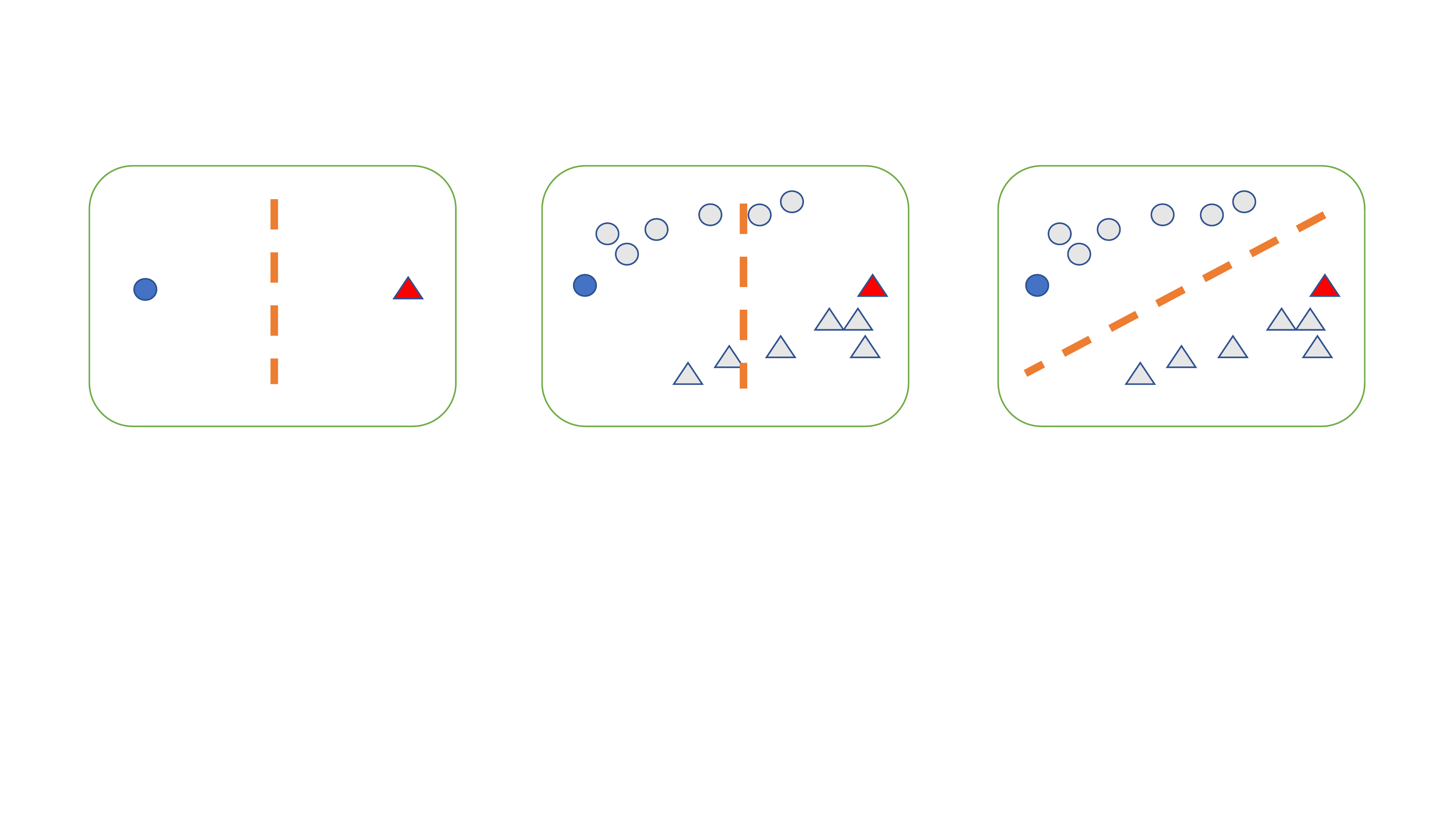}
    \vspace{-25mm}
	\caption{An example of the influence of decision boundary by unlabeled data. Blue circle point and red triangle point represent labeled dataset $S_L$. Grey points represent unlabeled dataset $U_D$. The shape of the points (circle or triangle) represents their true labels. The orange dashed line represents optimal decision boundary.}
	\label{SSL_benefits}
\end{figure}

Figure~\ref{SSL_benefits} shows a simple example to simulate and explain why RST is not an optimal solution. Following RST, the left panel shows the decision boundary (orange line) learning merely from the labeled dataset $S_L = \{$blue circle, red triangle$\}$. Based on $S_L$, the best annotator we get is the vertical orange line. It will perfectly classify the labeled dataset $S_L$ with zero errors. Nonetheless, it will inevitably annotate some of unlabeled data (grey points) with wrong labels (middle panel of Figure~\ref{SSL_benefits}). For example, if we adopted the orange line in left panel as the annotator, at least 2 grey circle points would be wrongly annotated as triangle and 2 grey triangle points wrongly annotated as circle.

To address above issues, we can train a classifier based on $S_L$ and $D_U$. Specifically, we first train a classifier based on $S_L$. Then, we use the pre-trained classifier to annotate all unlabeled data $D_U$, which acquires pseudo labels on unlabeled data to get $S_U$. We jointly use dataset $S = S_L \cup S_U$ to re-train the classifier, and then re-annotate $D_U$ via re-trained classifier until the convergence (i.e., \textit{multiple times}). Finally, we jointly use dataset $S = S_L \cup S_U$ to train a adversarially robust deep neural network. The key step is to use the re-trained classifier to annotate all unlabeled data $D_U$ in \textit{multiple times} during training.

Taking right panel of Figure~\ref{SSL_benefits} as an example, which learns a new decision boundary (i.e., a good annotator). This annotator utilizes labeled data $S_L$ together with unlabeled data (grey points), and these unlabeled data can elucidate the data distribution well.
%For example, the knowledge on $P(X)$ that one gains through the unlabeled data carry useful information in the inference of $P(Y|X)$. Thus, the new decision boundary (orange line) embedded the knowledge of both labeled data and unlabeled data, 
The annotator embedded with the knowledge of both labeled and unlabeled data can characterize the true distribution accurately. Thus, it will annotate ground-truth labels to those unlabeled data (grey points). To sum up, pre-determined annotator (left panel of Figure~\ref{SSL_benefits}) is not good enough to annotate unlabeled dataset $D_U$, which motivates us to explore re-trained annotator $f$ (Sections~\ref{simple-realization} and \ref{pow-realization}).

\subsection{The Simple Realization}\label{simple-realization}
The top simple realization is to utilize the conventional self-training~\cite{mcclosky2006effective_self_train,lee2013pseudo_self_train}. The key idea of self-training is to utilize DNN $f$'s predictions on unlabeled data over training iterations, namely annotating unlabeled data $D_U$ in \textit{multiple times}. Specifically, if the probability of $\mathbf{x}\in D_U$ assigned to the most likely class is higher than a predetermined threshold $\tau$, $\mathbf{x}$ is added to the training set for further training with $\Bar{y} = \arg\max f(\mathbf{x})$ as its pseudo label, i.e., $S_L \Leftarrow S_L \cup \{ (\mathbf{x}, \Bar{y})\}$. This process is repeated for a fixed number of iterations $T$ or until no more unlabeled data available or confident.

\begin{figure}[tp!]
	\centering
	\includegraphics[scale=0.22]{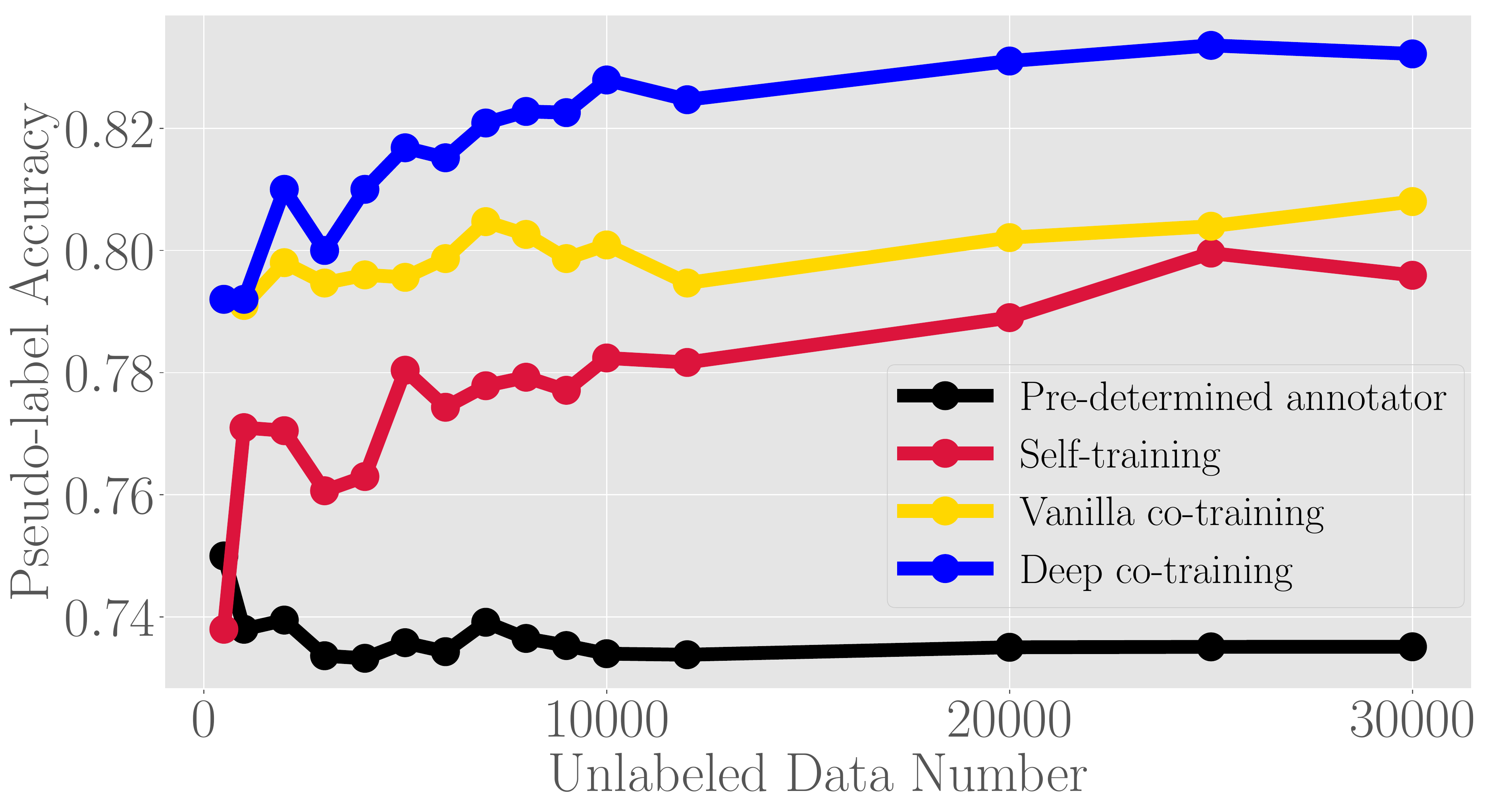}
	\caption{Comparisons on pseudo label accuracy on unlabeled data by the annotator obtained by various methods. Pre-determined annotator is used in RST~\cite{Percy}, self-training, vanilla co-training, deep co-training are our proposed methods to improve the quality of the annotator.}
	\label{boost_label_quality}
\end{figure}

Figure~\ref{boost_label_quality} empirically justifies the efficacy of self-training (red line), which significantly improves the quality of pseudo labels compared to pre-determined annotator (black line). However, there is a drawback in conventional self-training, namely network $f$ is hardly able to correct its own mistakes. Assume that the prediction of deep networks $f$ on an unlabeled data $\mathbf{x}$ is incorrect at the early training stage. Nonetheless, the data with incorrect pseudo label will be utilized in future training iterations. Due to memorization effect of deep networks~\cite{arpit2017closer}, $f$ will fit the wrongly-labeled data, which will hurt the test performance seriously~\cite{zhu2004class_noise}. This negative effects become even worse, when the domain of unlabeled data is different from that of labeled data~\cite{oliver2018realistic}.

To ameliorate the inferiority of self-training, the straightforward approach is to introduce a pair of networks correcting mistakes of each other, namely vanilla co-training~\cite{blum1998combining}. Specifically, we train a pair of DNNs (i.e., $f_1$ and $f_2$) simultaneously. We encourage two networks making consistent predictions on unlabeled data. Meanwhile, two DNNs are feed with different orders of labeled data to keep the inconsistent pace of training. To be specific, at each training iteration with $ (\mathbf{x}_1, y_1) \in \Bar{S}_1 \subseteq S_L$, $ (\mathbf{x}_2, y_2) \in \Bar{S}_2 \subseteq S_L$ and $\mathbf{x}_U \in \Bar{D}_U \subseteq D_U$, two deep networks $f_1$ and $f_2$ feed forward the common unlabeled data $\mathbf{x}_U \in D_U$ and different labeled data $\mathbf{x}_1$ and $\mathbf{x}_2$, and then update parameters $w_{f_1}$ and $w_{f_2}$ by
\begin{align}
     w_{f_1} = w_{f_1} - \eta \nabla  \displaystyle \big (    \mathcal{L}  (f_1(\mathbf{x}_1),y_1)  + \lambda   \mathcal{JS} (f_1(\mathbf{x}_U), f_2(\mathbf{x}_U) ) \displaystyle \big  ) \label{villa_co_train_eq1};\\
     w_{f_2} = w_{f_2} - \eta \nabla  \displaystyle \big (    \mathcal{L}  (f_1(\mathbf{x}_2),y_2)  + \lambda   \mathcal{JS} (f_2(\mathbf{x}_U), f_1(\mathbf{x}_U) ) \displaystyle \big  )\label{villa_co_train_eq2},
\end{align}
where $\eta$ is the learning rate, $\lambda$ is the trade-off parameter, $\mathcal{L}$ is cross entropy loss for labeled data, and $\mathcal{JS}$ is Jensen-Shannon divergence between two predicted probability between $f_1$ and $f_2$ on the same unlabeled data $\mathbf{x}_U$. 

We leverage Jensen-Shannon (JS) divergence to measure the similarity between two predicted probability between $f_1(\mathbf{x}_U)$ and $f_2(\mathbf{x}_U)$. The JS value is bounded and positive, and the smaller value denotes larger similarity between two probability distributions, vice versa. To minimize JS divergence between predicted probability between $f_1(\mathbf{x}_U)$ and $f_2(\mathbf{x}_U)$ on unlabeled data $\mathbf{x}_U$, Eq.~\eqref{villa_co_train_eq1} and \eqref{villa_co_train_eq2} encourage $f_1$ and $f_2$ making similar predictions on unlabeled data $\mathbf{x}_U$. Meanwhile, at each training iteration, two DNNs learn from different labeled data $(\mathbf{x}_1, y_1)$ and $(\mathbf{x}_2, y_2)$. This will keep each other diverged. Thus, two networks $f_1$ and $f_2$ could be complementary and could help to correct its peer's mistake on unlabeled data. Besides Jensen-Shannon divergence, we can also use other divergences, such as KL-divergence and Hellinger distance.

From Figure~\ref{SSL_benefits}, we observe an obvious improvement by vanilla co-training (yellow line) compared with self-training (red line). Taking a closer look at pseudo-label accuracy on unlabeled data, we find that the result of vanilla co-training is better than that of self-training. We believe that the interaction between peer networks (i.e., vanilla co-training) takes positive effects, while there is no any interaction in a single network (i.e., self-training). This point is also supported by the philosophy of collaborative learning~\cite{dillenbourg1999collaborative_learning}, where each member interacts with others actively by sharing experiences. Each member takes on asymmetric roles so that new knowledge can be created within members. Nonetheless, the improvement of label accuracy is not completely satisfying. When the number of unlabeled data increases from 20k to 30k, label accuracy of vanilla co-training is similar to that of self-training, since vanilla co-training has the collapsing problem. Namely, two networks gradually become the same one in function, which will not be able to correct mistakes of each other on unlabeled data.

As shown in Figure~\ref{total_variance} (yellow line), total variance of predictions $f_1(\mathbf{x})$ and $f_2(\mathbf{x})$ are large before 50 epochs. The high total variance denotes that two networks are diverged in function, since two networks have different views on unlabeled data at the initial training stage. Their different views come from different initialization and orders of fetching labeled data. The benefit of such divergence is that one network could have information gain from observing its peer network. Thus, they have sufficient capacities to correct mistakes of each other on unlabeled data. However, with the increase of training epochs, total variance of two networks gradually decreases and approaches near zero after 350 epochs. It means that two networks gradually converge to the same in function, and they can not correct mistakes of each other. Thus, vanilla co-training will gradually degenerate into self-training, which suffers from accumulated error problem at later training epochs.

%The loss to update the parameter
%Straightforward co-training have such problem that two network easily collapse into each other so that it degenerates quickly into self-training. 

\subsection{The Powerful Realization}\label{pow-realization}
To address the collapsing problem of vanilla co-training, we should keep two networks diverged in function. Especially at later training epochs, we should add an extra force pulling each other away, so that they always have capacities for correcting mistakes of each other. Inspired by~\cite{qiao2018deepcotraining}, we encourage two networks diverged by exploiting peer's adversarial examples, namely \textit{deep co-training}. 

In general, adversarial example is modified from natural example, where the network has large loss on adversarial example while it has small loss on its corresponding natural example. Adversarial example unveils the input space where the network could easily make mistakes. Such space is the weakest part (i.e., leading to unreliable prediction) corresponding to the network. Two networks can keep inconsistent from each other by learning from the weakest part of each other. Namely, each network robustly trains on adversarial examples generated by its peer network. Intuitively, each network always ``looks'' into peer's weakest part. Thus, two networks could prevent themselves from collapsing into one function and constantly keep diverged.

\newlength{\overwritelength}
\newlength{\minimumoverwritelength}
\setlength{\minimumoverwritelength}{1cm}
\newcommand{\overwrite}[3][red]{%
  \settowidth{\overwritelength}{$#2$}%
  \ifdim\overwritelength<\minimumoverwritelength%
    \setlength{\overwritelength}{\minimumoverwritelength}\fi%
  \stackrel
    {%
      \begin{minipage}{\overwritelength}%
        \color{#1}\centering\small #3\\%
        \rule{1pt}{9pt}%
      \end{minipage}}
    {\colorbox{#1!50}{\color{black}$\displaystyle#2$}}}

\begin{figure}[h!]
	\centering
	\includegraphics[scale=0.3]{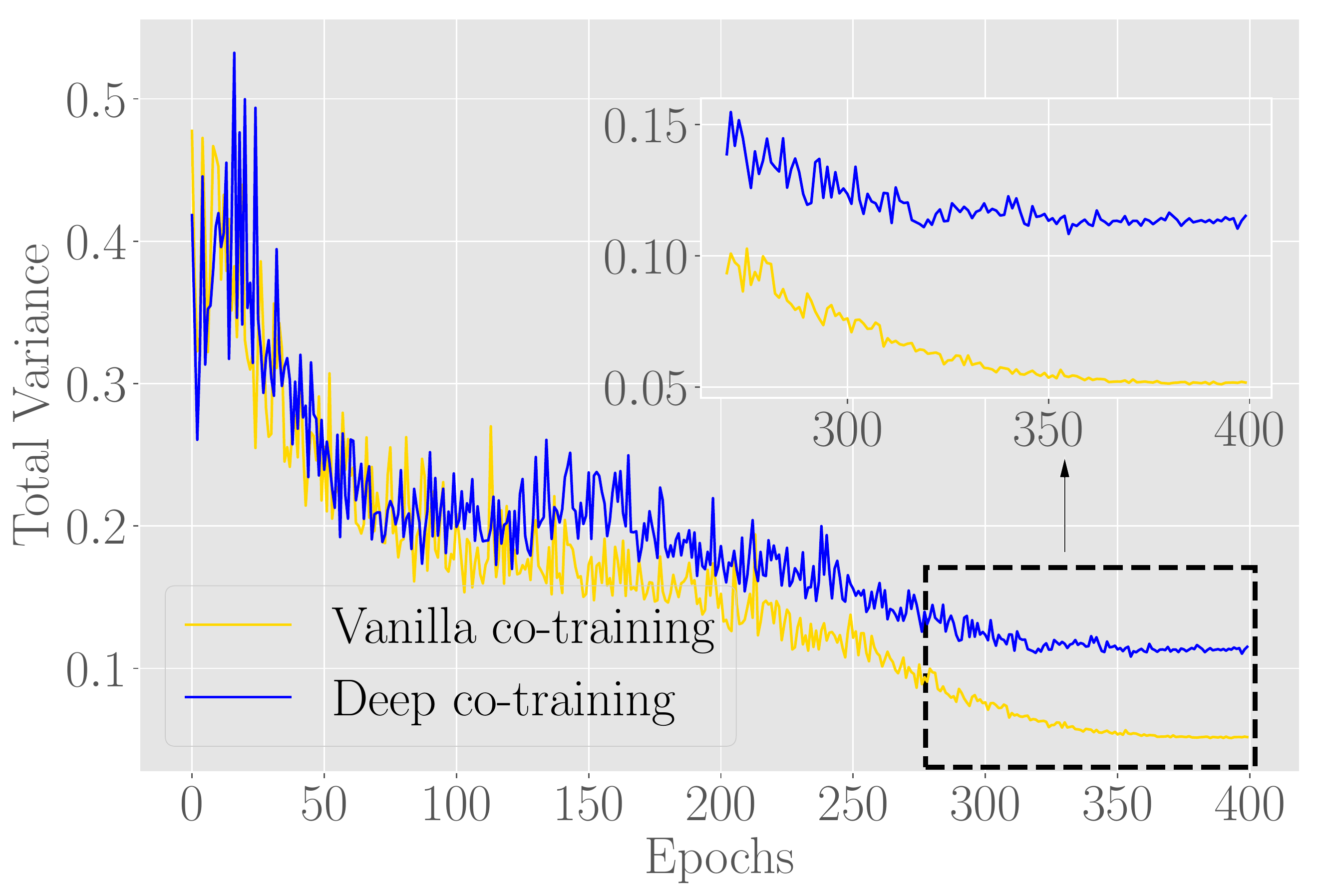}
	\caption{Comparison of divergence (evaluated by total variance) between two DNNs $f_1$ and $f_2$ trained by vanilla co-training and deep co-training. Total variance is empirically calculated by probability predictions by $f_1$ and $f_2$ on test data.}
	\label{total_variance}
\end{figure}

Mathematically, at each training iteration with $ (\mathbf{x}_1, y_1) \in \Bar{S}_1 \subseteq S_L$, $ (\mathbf{x}_2, y_2) \in \Bar{S}_2 \subseteq S_L$ and $\mathbf{x}_U \in \Bar{D}_U \subseteq D_U$,  two networks update themselves by
\begin{align}
    w_{f_1} = w_{f_1} &- \eta \nabla  \displaystyle \big (   \mathcal{L}  (f_1(\mathbf{x}_1),y_1)  \nonumber \\  
    &+ \lambda_1   \mathcal{JS} (f_1(\mathbf{x}_U), f_2(\mathbf{x}_U) )   \nonumber \\
 &+ \underbrace{  \lambda_2   \mathcal{JS} (f_1 (\mathbf{x}_{1}), f_2 (\mathbf{x}^{{adv}}_{1})  )}_{\text{Learn from $f_2$'s adversarial example of labeled data}} \nonumber \\ 
 &+ \underbrace{  \lambda_3 \mathcal{JS} (f_1 (\mathbf{x}_{U}), f_2 (\mathbf{x}^{{adv}_1}_U) )}_{\text{Learn from $f_2$'s adversarial example of unlabeled data}} 
 \displaystyle \big  );  \label{co_adversarial_train_eq1} \\ 
    w_{f_2} = w_{f_2} &- \eta \nabla  \displaystyle \big (   \mathcal{L}  (f_2(\mathbf{x}_2),y_2)  \nonumber \\  
    &+ \lambda_1   \mathcal{JS} (f_2(\mathbf{x}_U), f_1(\mathbf{x}_U) )   \nonumber \\
 &+ \underbrace{ \lambda_2  \mathcal{JS} (f_2 (\mathbf{x}_{2}), f_1 (\mathbf{x}^{{adv}}_{2})  ) }_{\text{Learn from $f_1$'s adversarial example of labeled data}}  \nonumber \\
 &+ \underbrace{ \lambda_3 \mathcal{JS} (f_2 (\mathbf{x}_{U}), f_1 (\mathbf{x}^{{adv}_2}_{U}) ) }_{\text{Learn from $f_1$'s adversarial example of unlabeled data}} 
 \displaystyle \big  ), \label{co_adversarial_train_eq2}
\end{align}

where $w_{f_1}$ and $w_{f_2}$ are weights of $f_1$ and $f_2$, $\eta$ is the learning rate, $\lambda_1$, $\lambda_2$ and $\lambda_3$ are trade-off parameters, $\mathcal{L}$ is cross entropy loss for labeled data, and $\mathcal{JS}$ is Jensen-Shannon divergence between two predicted probability between $f_1$ and $f_2$. Most importantly, adversarial data $\mathbf{x}^{{f_2}_{adv}}_{1}$, $ \mathbf{x}^{{f_1}_{adv}}_{2}$, $\mathbf{x}^{{f_2}_{adv}}_{U}$ and $\mathbf{x}^{{f_1}_{adv}}_{U}$ are exploited due to 

\begin{align}
\mathbf{x}^{{adv}}_{1} &= \displaystyle \underset{|| \mathbf{x}^{{adv}}_{1} -  \mathbf{x}_{1} || \leq \epsilon}{\arg\max} \mathcal{L} (f_1( \mathbf{x}^{{adv}}_{1}) , y_1); \label{S_1_adv_on_f1}\\ 
\mathbf{x}^{{adv}}_{2} &=\displaystyle \underset{|| \mathbf{x}^{{adv}}_{2} -  \mathbf{x}_{2} || \leq \epsilon}{\arg\max} \mathcal{L} (f_2( \mathbf{x}^{{adv}}_{2}) , y_2); \label{S_2_adv_on_f2}\\
\mathbf{x}^{{adv}_1}_{U} &= \displaystyle \underset{|| \mathbf{x}^{{adv}_1}_{U} -  \mathbf{x}_{U} || \leq \epsilon}{\arg\max} \mathcal{L} (f_1( \mathbf{x}^{{adv}_1}_{U}) , y_{U}^{f_1});\label{U_adv_on_f1} \\
\mathbf{x}^{{adv}_2}_{U} &= \displaystyle \underset{|| \mathbf{x}^{{adv}_2}_{U} -  \mathbf{x}_{U} || \leq \epsilon}{\arg\max} \mathcal{L} (f_2( \mathbf{x}^{{adv}_2}_{U}) , y_{U}^{f_2}),\label{U_adv_on_f2}
\end{align}
where $ y_{U}^{f_1}$, $ y_{U}^{f_2}$ are predicted labels by networks $f_1$ and $f_2$ on $\mathbf{x}_U$ respectively, and $\mathcal{L}$ is cross entropy loss.

Compared with vanilla co-training, the extra loss terms under-braced in Eq. \eqref{co_adversarial_train_eq1} and \eqref{co_adversarial_train_eq2} are introduced in the deep co-training. Note that $\lambda_2$, $\lambda_3$ in Eq. \eqref{co_adversarial_train_eq1} and \eqref{co_adversarial_train_eq2} and $\epsilon$ in Eq. \eqref{S_1_adv_on_f1} - \eqref{U_adv_on_f2} control the ``force'' that pulls each other away. Specifically, $\lambda_2$ and $\lambda_3$ control importance of divergence term under-braced, while $\epsilon$ decides allowable size of norm ball around the natural data, where adversarial examples are generated. Increasing $\lambda_2$, $\lambda_3$ and $\epsilon$ could enable more divergence between two networks. In practice, Eq. \eqref{S_1_adv_on_f1} - \eqref{U_adv_on_f2} are hard to be solved analytically, and thus we approximate its solution by PGD~\cite{MadryMSTV18} or FGSM~\cite{goodfellow_FGSM}. 

Figure~\ref{boost_label_quality} validates the efficacy of deep co-training (blue line). We set $\lambda_1 = 10.0$, $\lambda_2 = \lambda_3 = 0.5$, and $\epsilon=0.02$ in Eq. \eqref{co_adversarial_train_eq1} and \eqref{co_adversarial_train_eq2}. We utilize FGSM with single step to search for peer network's adversarial example. For fair comparison, we set $\lambda = 10.0$ in Eq. \eqref{villa_co_train_eq1} and \eqref{villa_co_train_eq2} (vanilla co-training). It empirically shows that the quality of pseudo labels by deep co-training (blue line) is significantly higher than that of vanilla co-training (yellow line).

To deeply understand the deep co-training, we analyze total variance of two networks over training epochs (blue line in Figure~\ref{total_variance}). Similar to vanilla co-training, both networks start to converge to each other. However, at late training epochs (e.g., after 300 epochs), total variance of vanilla co-training will approach near zero. In contrast, total variance of deep co-training can keep a positive value, since deep co-training exploits peer's adversarial examples and prevents two networks collapsing into the same in function. This brings us Algorithm~\ref{robust_co_adversarial_training} called robust co-training, which connects the deep co-training and adversarial training. Our proposed algorithm can empirically boost both standard accuracy and robust accuracy as follows.
%and what benefits? See Algorithm 1.
\SetKwInput{KwInput}{Input}                % Set the Input
\SetKwInput{KwOutput}{Output}
\SetKwInput{KwFetch}{Fetch}
\SetKwInput{KwObtain}{Obtain}
\SetKwInput{KwUpdate}{Update}
\SetKwInput{KwLabel}{Label}

\begin{algorithm}

\label{robust_co_adversarial_training}
\DontPrintSemicolon
  \KwInput{ $w_{f_1}$, $w_{f_2}$, $\theta_g$, maximum training epoch $T_L$, learning rate $\eta(T)$, trade off hyperparameters $\lambda_1(T)$ and $\lambda_2(T)$
  
  \KwData{Labeled dataset $S_L = \{ (\mathbf{x}_i, y_i ) \}^{N_L-1}_{i=0}$}
%  $\sum_{i=1}^{\infty} := 0$ %\tcp*{this is a comment}
%  
\KwFetch{Unlabeled dataset $D_U = \{\mathbf{x}_j \}^{N_U-1}_{j = 0 }$}}
    \tcc{Deep co-training}
    \For{$T$ = 1,2,..., $T_L$}
    { 
    \KwFetch{Mini-batch $\Bar{S}_1$ and $\Bar{S}_2$ from $S_L$, $\Bar{D}$ from $D_U$, the set of input domain of $\Bar{S}_1$ and $\Bar{S}_2$ is denoted as $\Bar{D}_{S_1}$,$\Bar{D}_{S_2}$.} 
    \KwObtain{Adversarial set $\Tilde{S}_1$, $\Tilde{D}_{1}$ targeting on $f_1$ on labeled dataset $\Bar{S}_1$ and unlabeled dataset $\Bar{D}$ according to Eq.~\eqref{S_1_adv_on_f1} and Eq.~\eqref{U_adv_on_f1}. \tcp*{$| \Tilde{S}_{1}| = |\Bar{S}_1| $, $|\Tilde{D}_{1}| =  |\Bar{D}|$} }  
    \KwObtain{Adversarial set $\Tilde{S}_2$, $\Tilde{D}_{2}$ targeting on $f_2$ on labeled set $\Bar{S}_2$ and unlabeled set $\Bar{D}$ according to Eq.~\eqref{S_2_adv_on_f2} and Eq.~\eqref{U_adv_on_f2}. \tcp*{$|\Tilde{S}_2| = |\Bar{S}_2|$, $| \Tilde{D}_{2}| = |\Bar{D}|$} }  
 
    \KwUpdate{ $w_{f_1} = w_{f_1} - \eta \nabla  \displaystyle \big ( \displaystyle   \sum_{(\mathbf{x},y) \in \Bar{S}_1} \mathcal{L}  (f_1(\mathbf{x}),y)  +
    \lambda_{1} \displaystyle  \sum_{\mathbf{x} \in \Bar{D}} \mathcal{JS} (f_1(\mathbf{x}), f_2(\mathbf{x})) + 
    \lambda_{2}  \displaystyle  \sum_{\mathbf{x} \in \Bar{D}_{S_1},  \mathbf{x}' \in \Tilde{S}_1} \mathcal{JS} (f_1(\mathbf{x}) , f_2 (\mathbf{x}')  + 
    \lambda_{3}  \displaystyle  \sum_{\mathbf{x} \in \Bar{D},  \mathbf{x}' \in \Tilde{D}_1} \mathcal{JS} (f_1(\mathbf{x}) , f_2 (\mathbf{x}')
    \displaystyle \big ) $  } 
    \KwUpdate{ $w_{f_2} = w_{f_2} - \eta \nabla  \displaystyle \big ( \displaystyle   \sum_{(\mathbf{x},y) \in \Bar{S}_2} \mathcal{L}  (f_2(\mathbf{x}),y)  +
    \lambda_{1} \displaystyle  \sum_{\mathbf{x} \in \Bar{D}} \mathcal{JS} (f_2(\mathbf{x}), f_1(\mathbf{x})) + 
    \lambda_{2}  \displaystyle  \sum_{\mathbf{x} \in \Bar{D}_{S_2}, \mathbf{x}' \in \Tilde{S}_2} \mathcal{JS} (f_2(\mathbf{x}) , f_1 (\mathbf{x}') + 
    \lambda_{3}  \displaystyle  \sum_{\mathbf{x} \in \Bar{D}, \mathbf{x}' \in \Tilde{D}_2} \mathcal{JS} (f_2(\mathbf{x}) , f_1 (\mathbf{x}') \displaystyle \big ) $  } 
    %\tcp*{$\mathcal{L} $ is cross entropy loss over labeled dataset and $JS$ is Jason divergence on output logit between two models. }
    }
    \KwLabel {Annotate $D_U$ using either $f_1$ or $f_2$ to get $S_U$.}
    \KwObtain{Augmented dataset $S = S_L \cup S_U$.}
    \tcc{Adversarial training}
    $   \displaystyle \theta_g =
   \underset{\theta_g}{\arg\min}   \frac{1}{|S|}  \sum_{x,y \in S } \displaystyle [ \mathcal{L}  (  g_{\theta}(\mathbf{x}), y) \nonumber +  \underset{|| \mathbf{x}' - \mathbf{x}|| \leq \epsilon }{max}
    \lambda \mathcal{KL} (g_{\theta}(\mathbf{x}),  g_{\theta}(\mathbf{x}')) \displaystyle ]$ 
    \tcp*{Use  Eq.(\ref{madry_adversarial_train}) or Eq.(\ref{trades}) }
    \KwOutput{$\theta_g$}
\caption{Robust co-training (RCT)}
\end{algorithm}
\begin{remark}
In Algorithm~\ref{robust_co_adversarial_training}, the quality of pseudo labels on unlabeled data get improved significantly via deep co-training. Thus, we could obtain augmented dataset $S$ by joining labeled dataset $S_L$ and high-quality pseudo-labeled dataset $S_U$. Then, we train a adversarially robust deep network on $S$ using either Madry's adversarial training (i.e., Eq. \eqref{madry_adversarial_train}) \cite{MadryMSTV18} or TRADES (i.e., Eq. \eqref{trades}) \cite{Zhang_trades}.
\end{remark}

\section{Experiments}
%\footnote{+++ Please add shade (variance) with three times run.}
%Baselines:
%1) random labels; 2) self-training; \footnote{+++ should be self-training + single step (Percy); self-training + multiple steps (ours);} 3) straightforward Co-training (ours); 4) Our proposed Co-adversarial training (ours). By the way, you should also add Oracle clean labels as a upper bound.
%\footnote{+++ I don't know why you mention this sentence, and no much use. JF's ANS: deleted.}. 
We conduct experiments on real-world dataset \textit{CIFAR-10} and \textit{SVHN}~\cite{netzer2011reading_SVHN}. We make comparisons between our robust co-training (RCT) and robust self-training (RST)~\cite{Percy}. We show our algorithm could give better pseudo label than RST. As a result, our algorithm boosts both standard test accuracy and robust test accuracy of adversarial training by a large margin.  
Thus, we empirically justify our main claim: The quality improvement of pseudo labels on unlabeled data could lead to the better adversarial training.
% \footnote{+++ why only compare with RST? why not vanilla Co-training?. JF's ANS: The logic line is: 1. Improvement of label quality can lead to improvement of adversarial training (from our motivation figure). 2. We need to find a method to improve label accuracy 3. We have three candidate, we choose a best one. 4. Thus, we only use the best one to improve the label quality. 5. Then The best one lead to better adversarial training. 6. Therefore, we only compare co adversarial robust training and original Percy's robust self-training.}

\subsection{Quality Improvement of Pseudo Labels}
Deep co-training in Algorithm~\ref{robust_co_adversarial_training} could achieve a significant improvement on pseudo-label accuracy. Compared to pre-determined annotator used in RST, we could annotate unlabeled data more accurately. Especially, when there are more unlabeled data available, deep co-training could further increase the quality of pseudo labels while pre-determined annotator do not have such effects.
\begin{figure}[h!]
	\centering
	\includegraphics[scale=0.20]{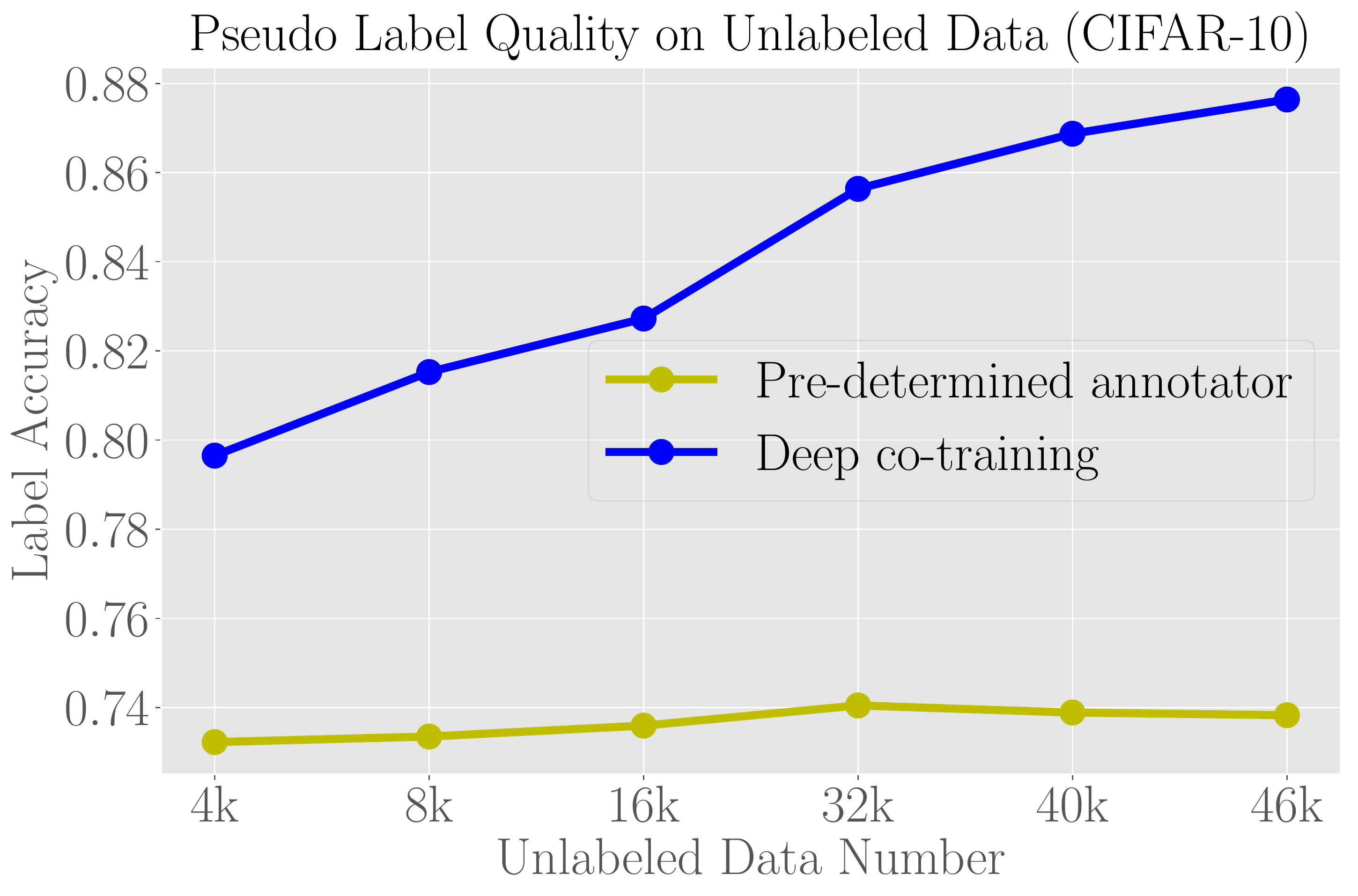}
	\includegraphics[scale=0.20]{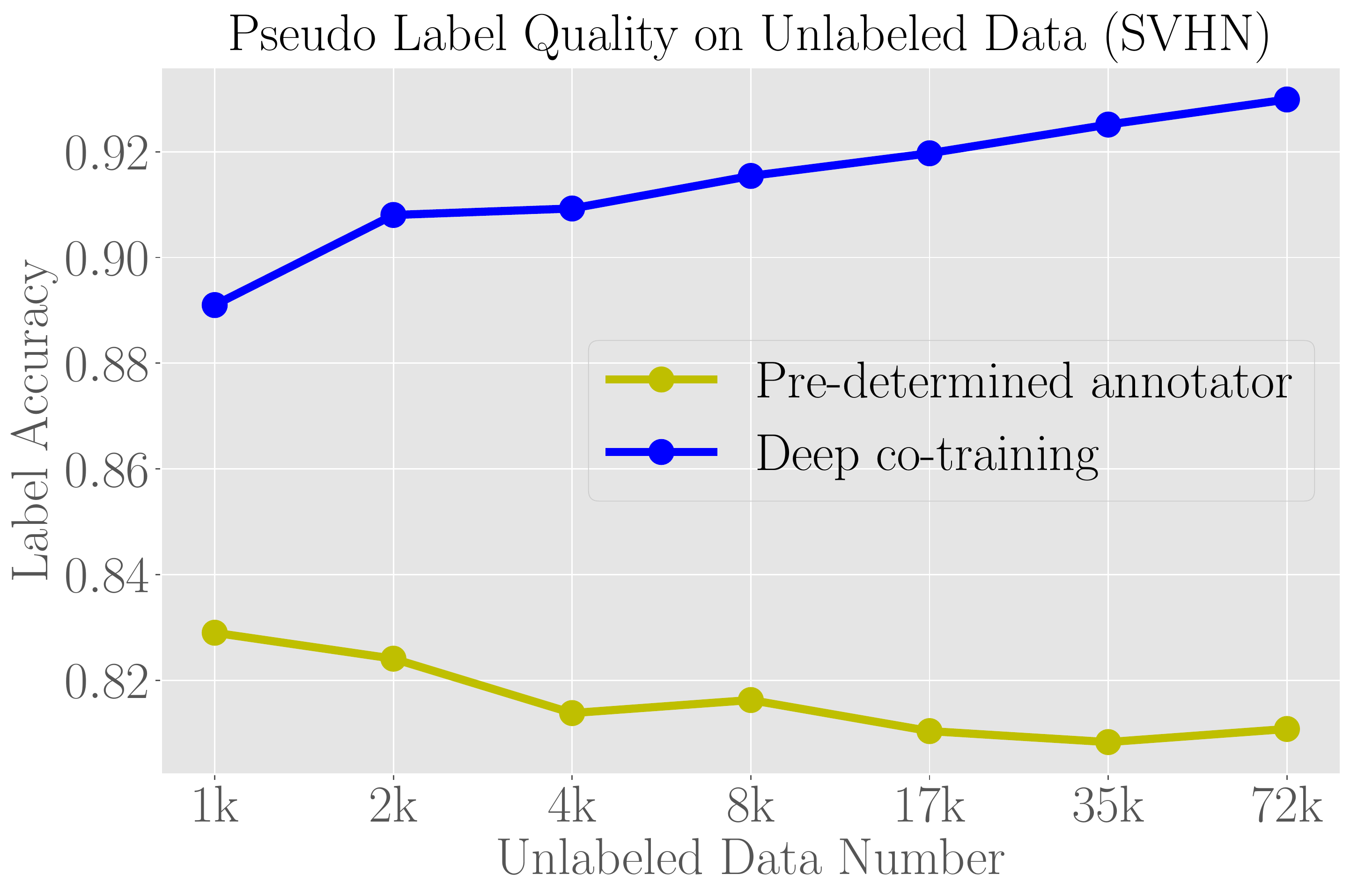}
	\caption{Pseudo label quality comparisons between using pre-determined annotator and deep co-training. Pre-determined annotator trained based on a single CNN13. Deep co-training are based on a pair of CNN13.}
	\label{label_quality_compara}
\end{figure}
%\footnote{+++ I just see CNN, not ResNet here, also, your y-axis is label correction, not consistent with previous label accuracy JF's ANS: corrected!}

To justify these effects, we randomly select 4k training data as labeled set $S_L$ and simulate the remaining 4k, 8k, 16k, 32k, 40k, 46k unlabeled dataset $D_U$ in \textit{CIFAR-10} dataset. In \textit{SVHN} dataset, we randomly select 1k training data as labeled dataset $S_L$ and simulate the remaining 1k, 2k, 4k, 8k, 17k, 35k, 72k as unlabeled dataset $D_U$. 

In Figure~\ref{label_quality_compara}, we compare pseudo-label accuracy on unlabeled data generated by pre-determined annotator~\cite{Percy} and deep co-training, respectively. We use CNN13~\cite{Laine_TemporalEnsembling} as the network backbone. Specifically, pre-determined annotator utilizes only a single CNN13, and deep co-training training utilizes a pair of CNN13.
% \footnote{+++ Why not ResNet? or WideResNet? How percy's paper does?JF's ANS: Percy's paper used tiny 80 million dataset and WideResNet.} 
For pre-determined annotator, we train a single CNN13 based on merely $S_L$ until convergence. Then we use the converged CNN13 to yield pseudo labels on all unlabeled data (yellow line).

In deep co-training, we learn a annotator involving unlabeled data. During training, we keep two networks diverged in function by setting $\lambda_1 = 10$ and $\lambda_2 = \lambda_3 = 0.5$ in Eq.~\eqref{co_adversarial_train_eq1} and Eq.~\eqref{co_adversarial_train_eq2}, which is inspired by \cite{qiao2018deepcotraining}. We co-train two CNN13 according to the Algorithm~\ref{robust_co_adversarial_training}, where maximum epoch $T_L = 600$, SGD with 0.9 momentum and learning rate $\eta$ starting from $0.1$ and decaying over epochs. Adversarial examples of Eq. \eqref{S_1_adv_on_f1} - \eqref{U_adv_on_f2} are generated by FGSM~\cite{goodfellow_FGSM} with single step, and the $\epsilon$ is set to $0.02$. Then, we randomly choose one of CNN13 pair as the annotator to label the unlabeled data.

Figure~\ref{label_quality_compara} shows the gap of pseudo-label accuracy between pre-determined annotator and deep co-training in both \textit{CIFAR-10} and \textit{SVHN} dataset. Specifically, pre-determined annotator (yellow line) does not involve unlabeled data in the learning process. When there are more unlabeled data available, pseudo-label accuracy on unlabeled data does not increase and even decrease. Therefore, RST~\cite{Percy} leveraging pre-determined annotator does not incorporate the knowledge of unlabeled data, and performs undesirably (Section \ref{improved-perf-adv}). By comparison, deep co-training (blue line) incorporates unlabeled data to learn the annotator. Besides, it introduces paradigm of collaborative learning to correct mistakes of each other. As a results, when there are more unlabeled data available, pseudo-label accuracy will increase correspondingly. Therefore, RCT (Algorithm~\ref{robust_co_adversarial_training}) leveraging re-trained annotator incorporates the knowledge of unlabeled data, and performs desirably (Section \ref{improved-perf-adv}).

\subsection{Improved Performance of Adversarial Training}\label{improved-perf-adv}

\begin{figure}[h!]

	\centering
	\includegraphics[scale=0.12]{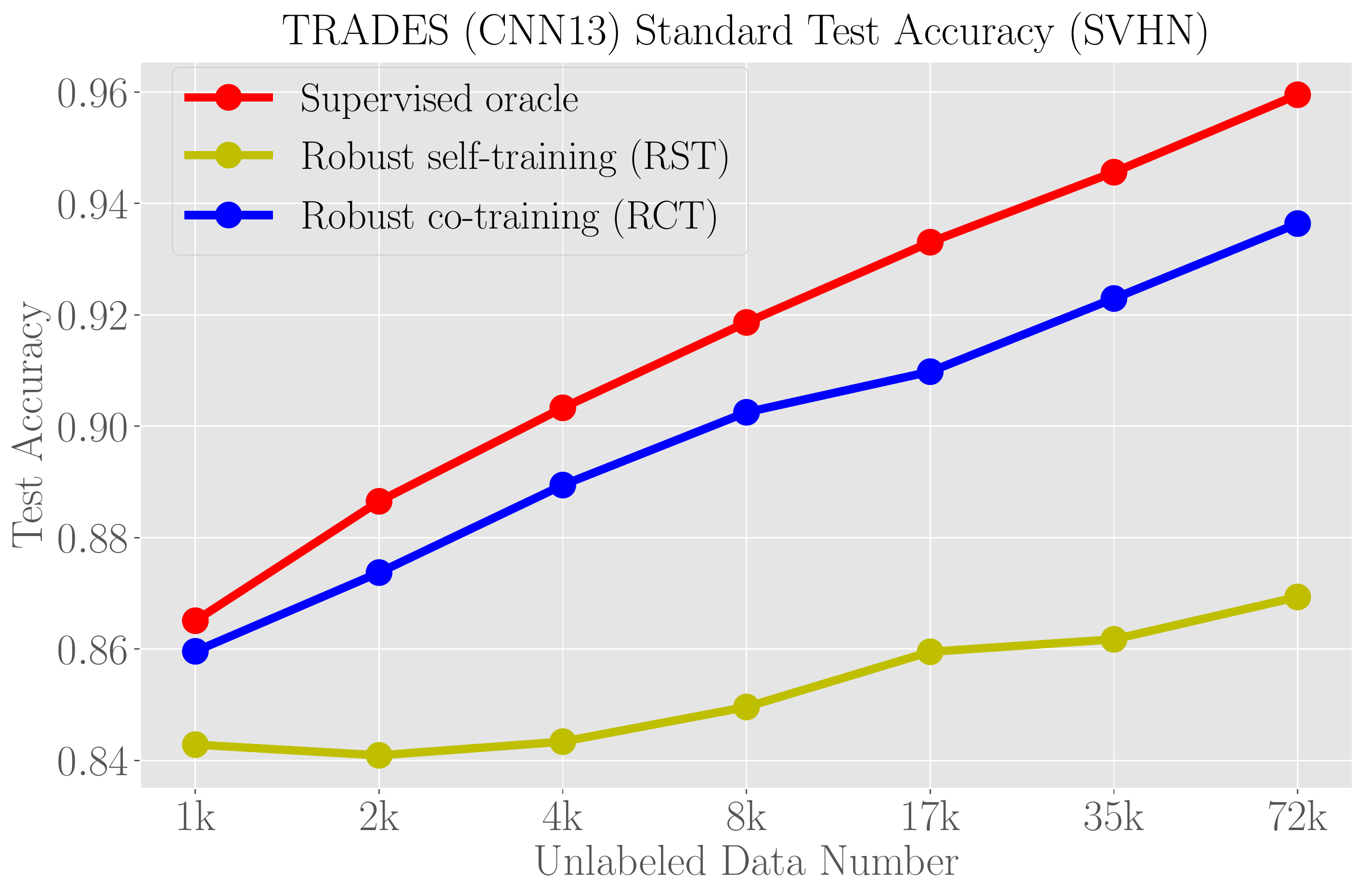}  
	\includegraphics[scale=0.12]{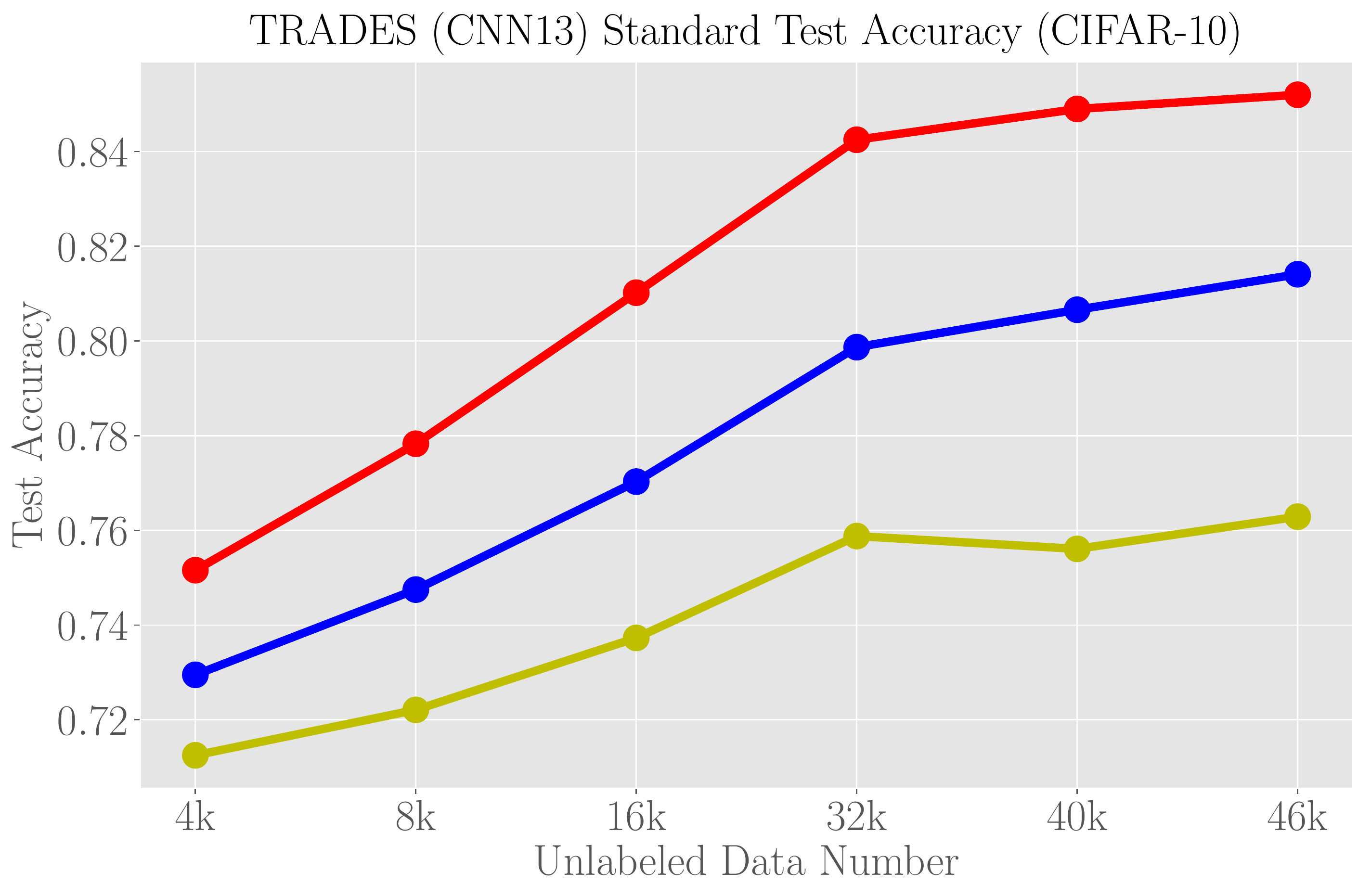} \includegraphics[scale=0.12]{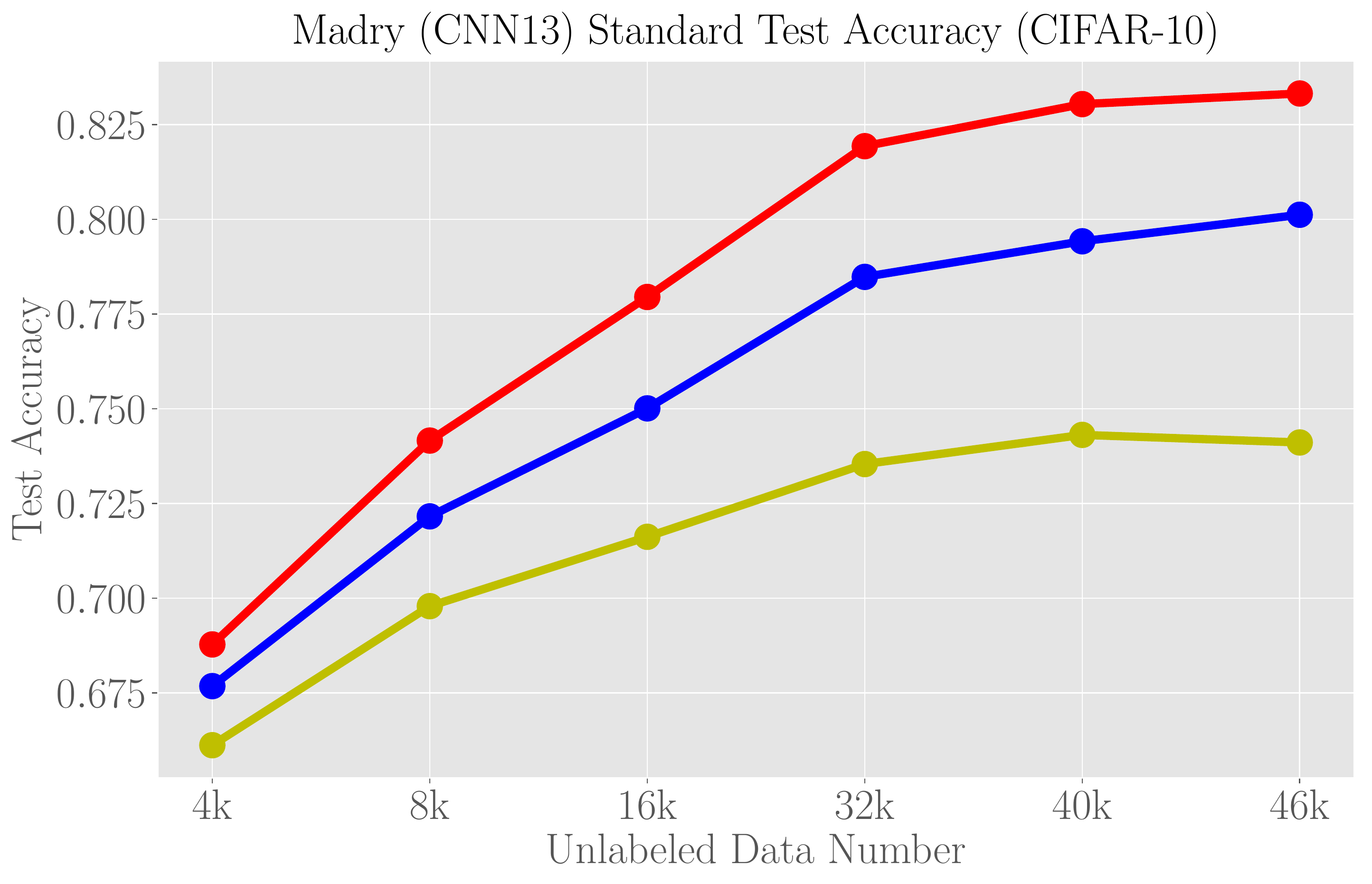} 
    \includegraphics[scale=0.12]{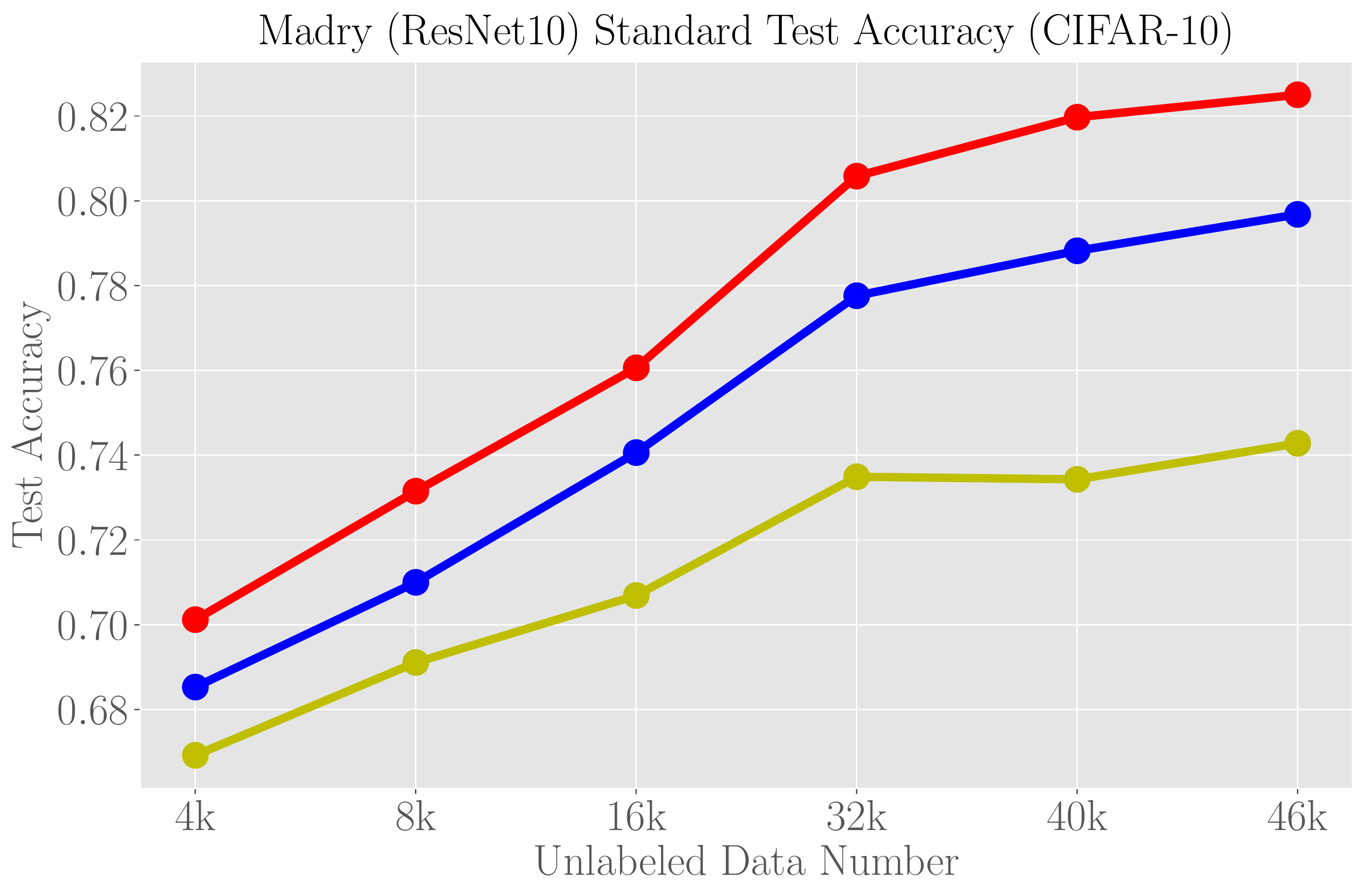} \\
	\includegraphics[scale=0.12]{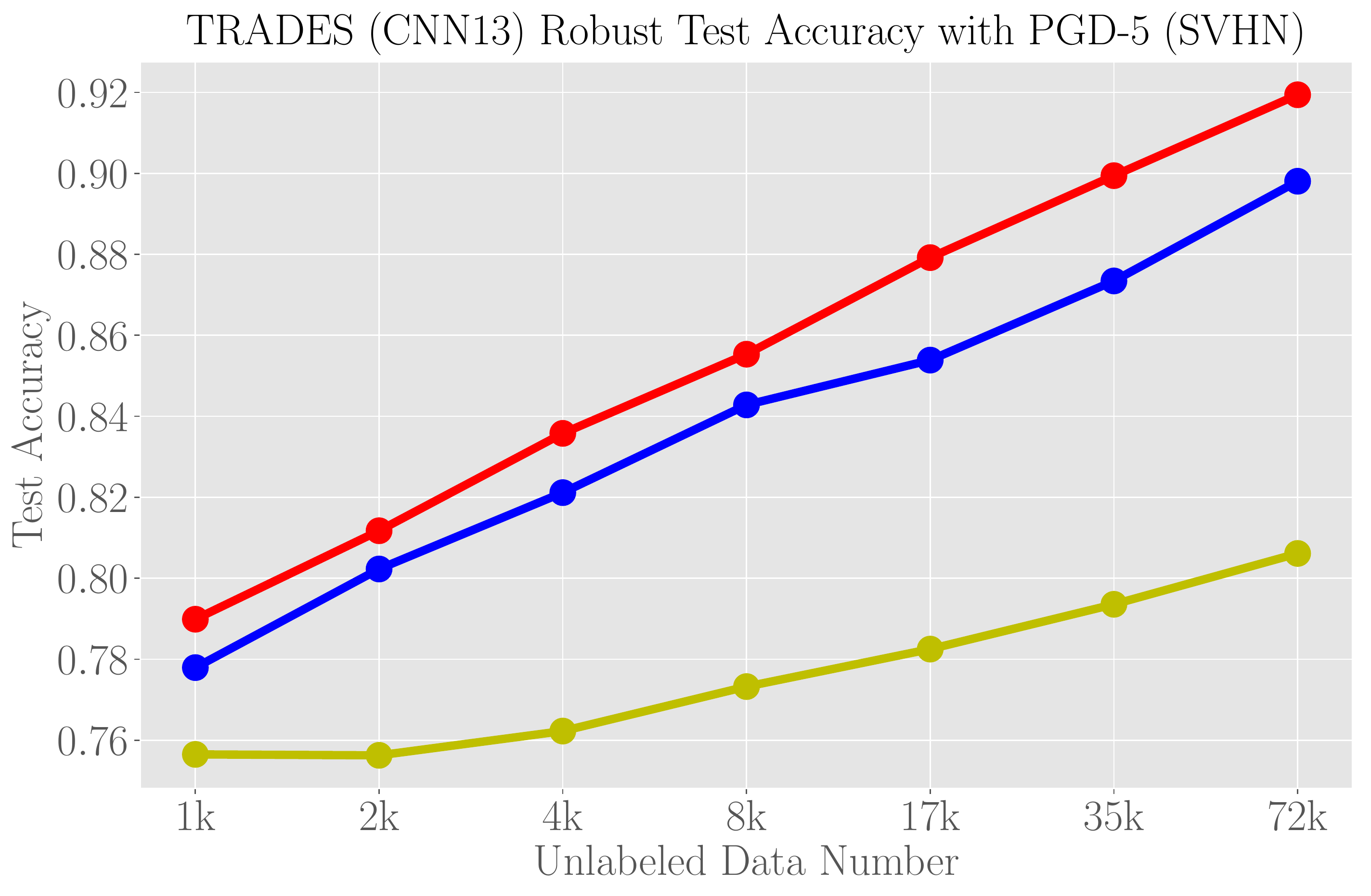}  
	\includegraphics[scale=0.12]{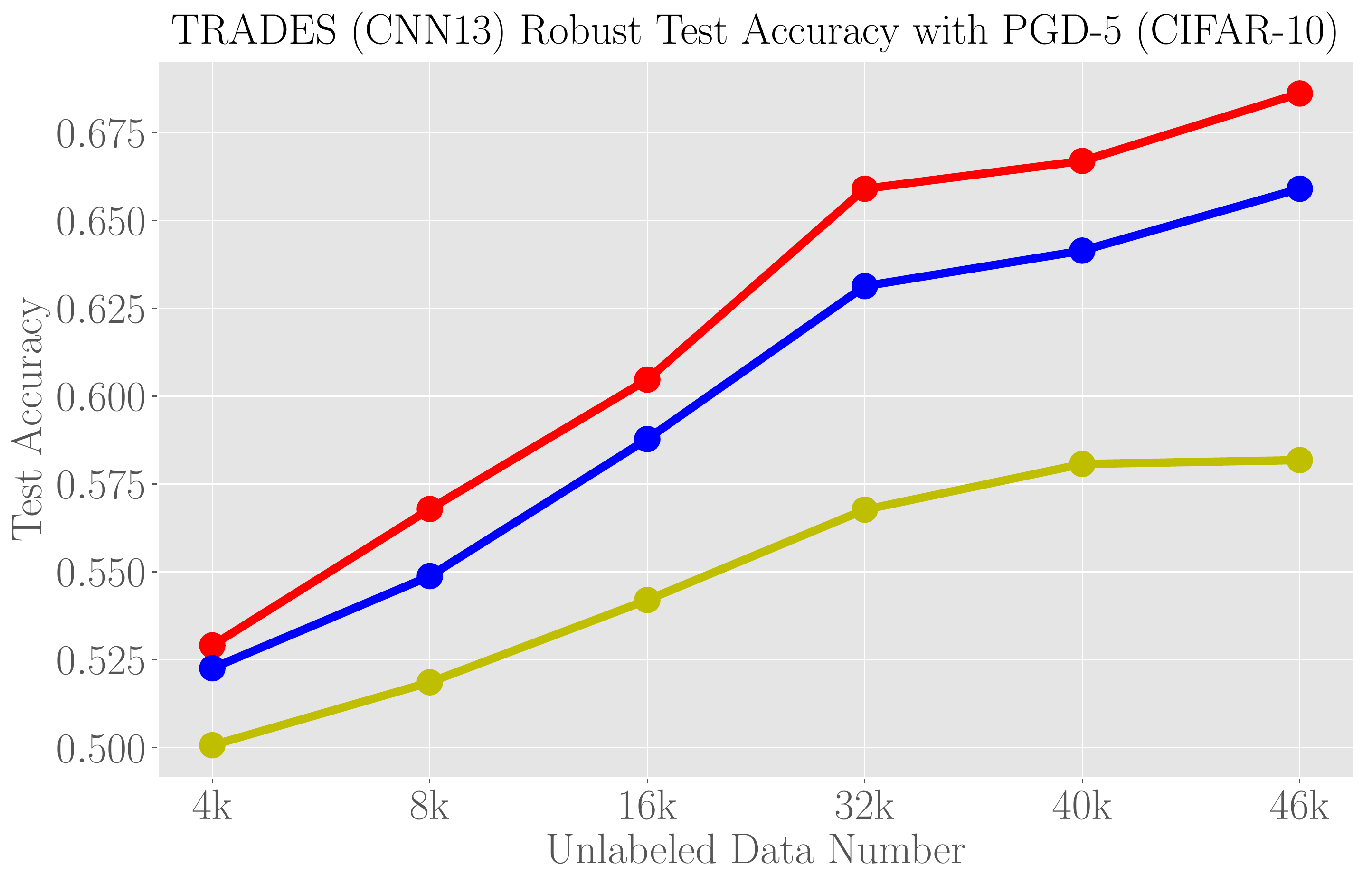} \includegraphics[scale=0.12]{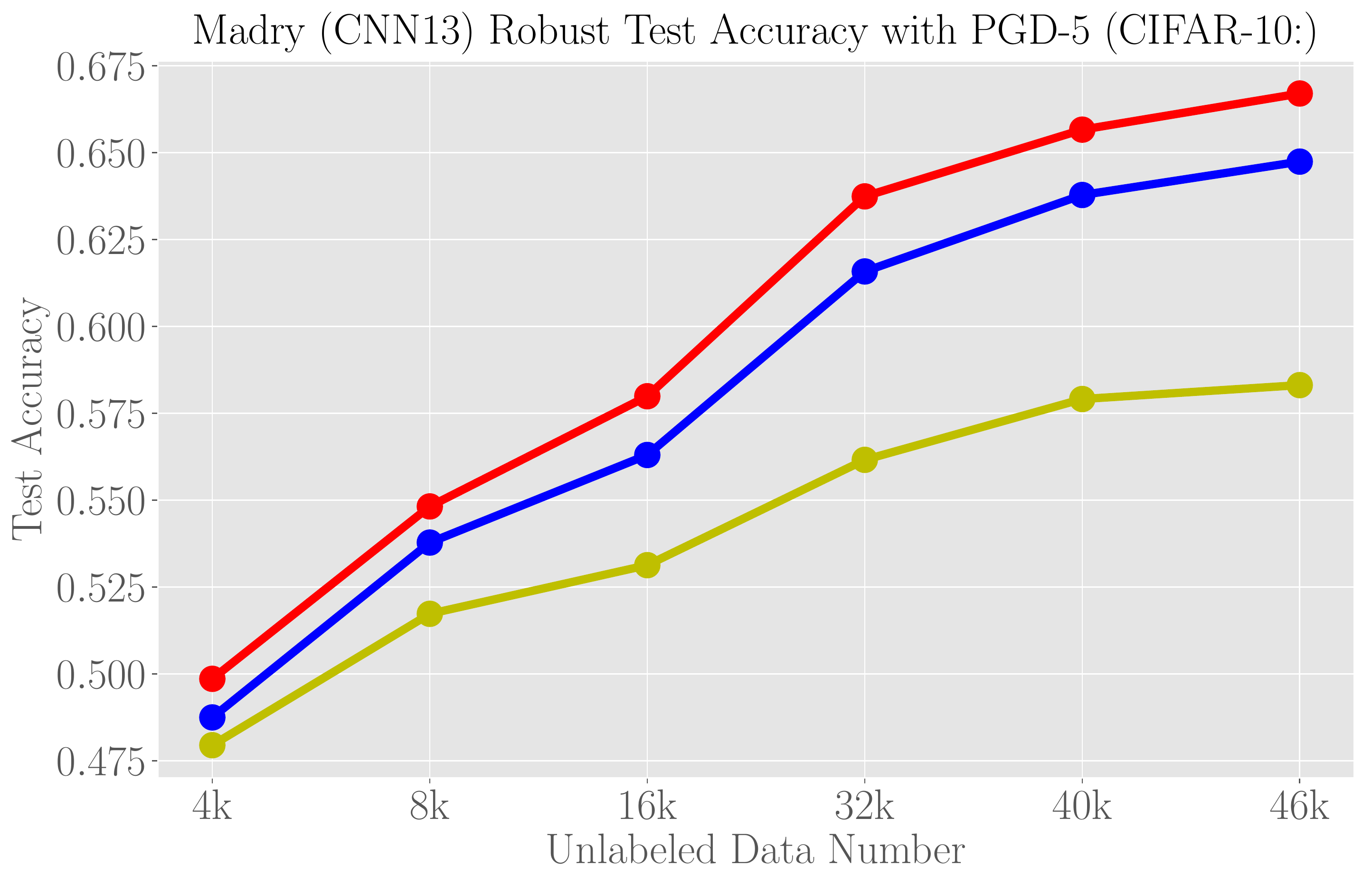} 
    \includegraphics[scale=0.12]{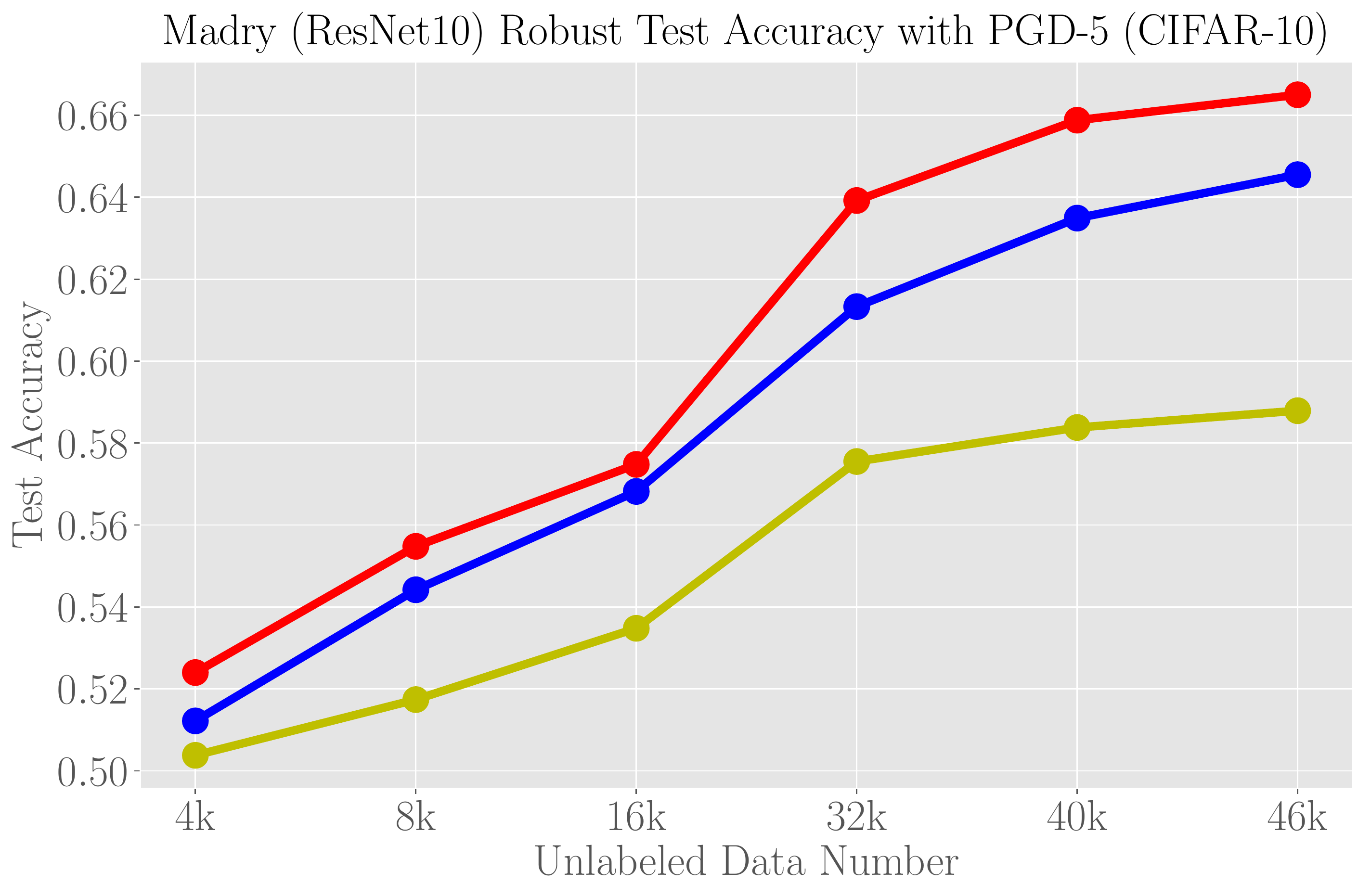} \\
	\includegraphics[scale=0.12]{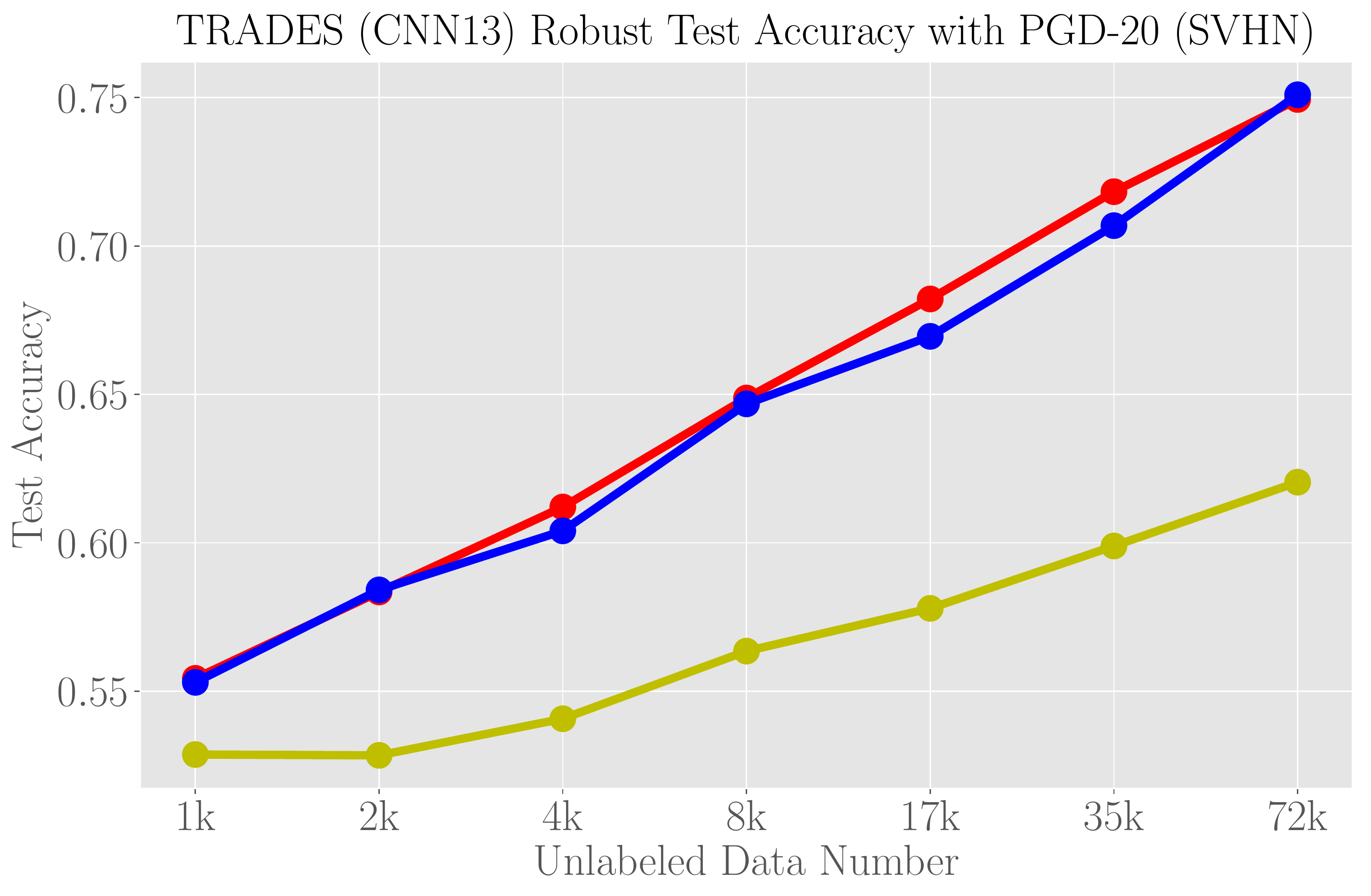}  
	\includegraphics[scale=0.12]{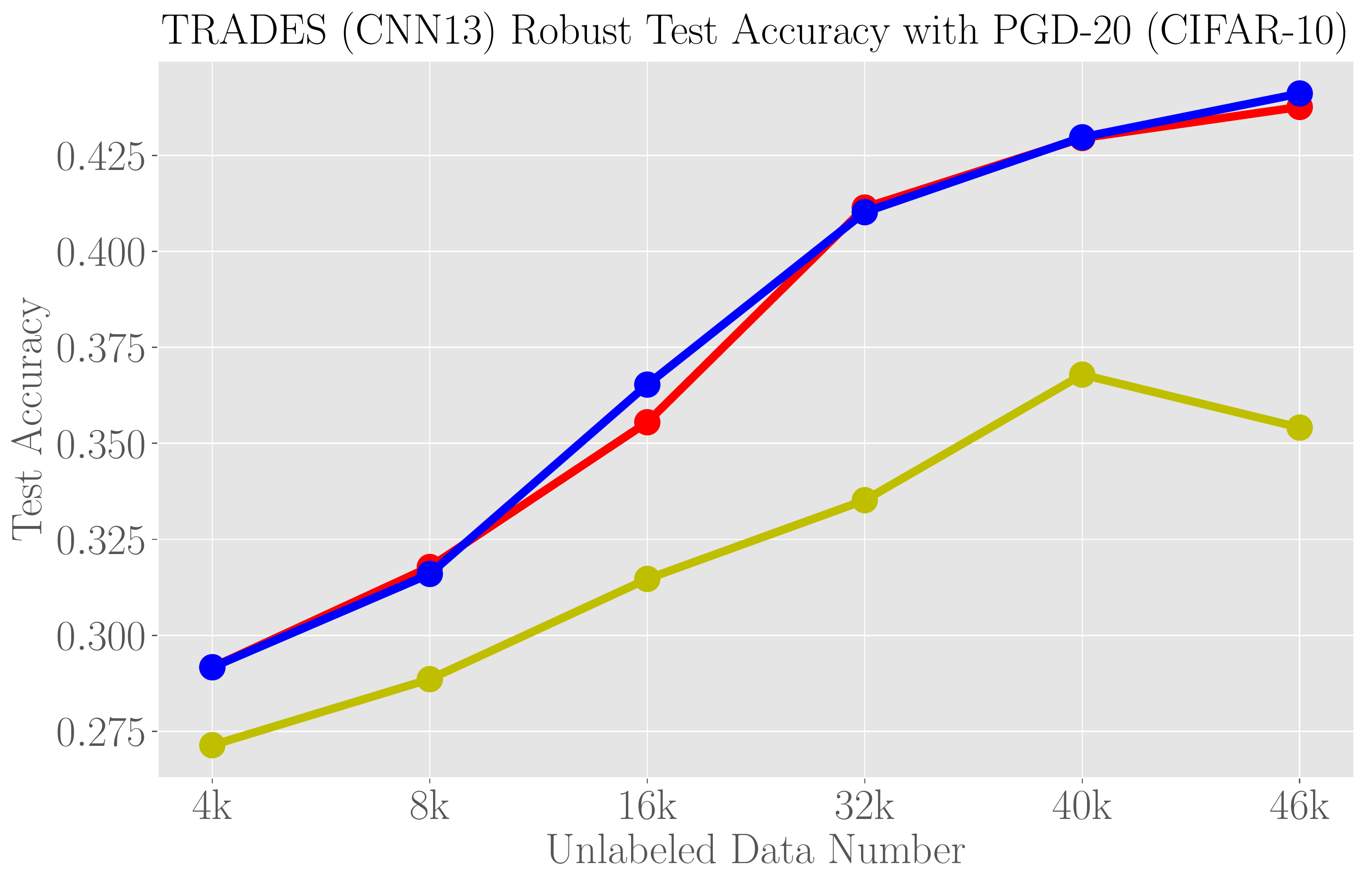} \includegraphics[scale=0.12]{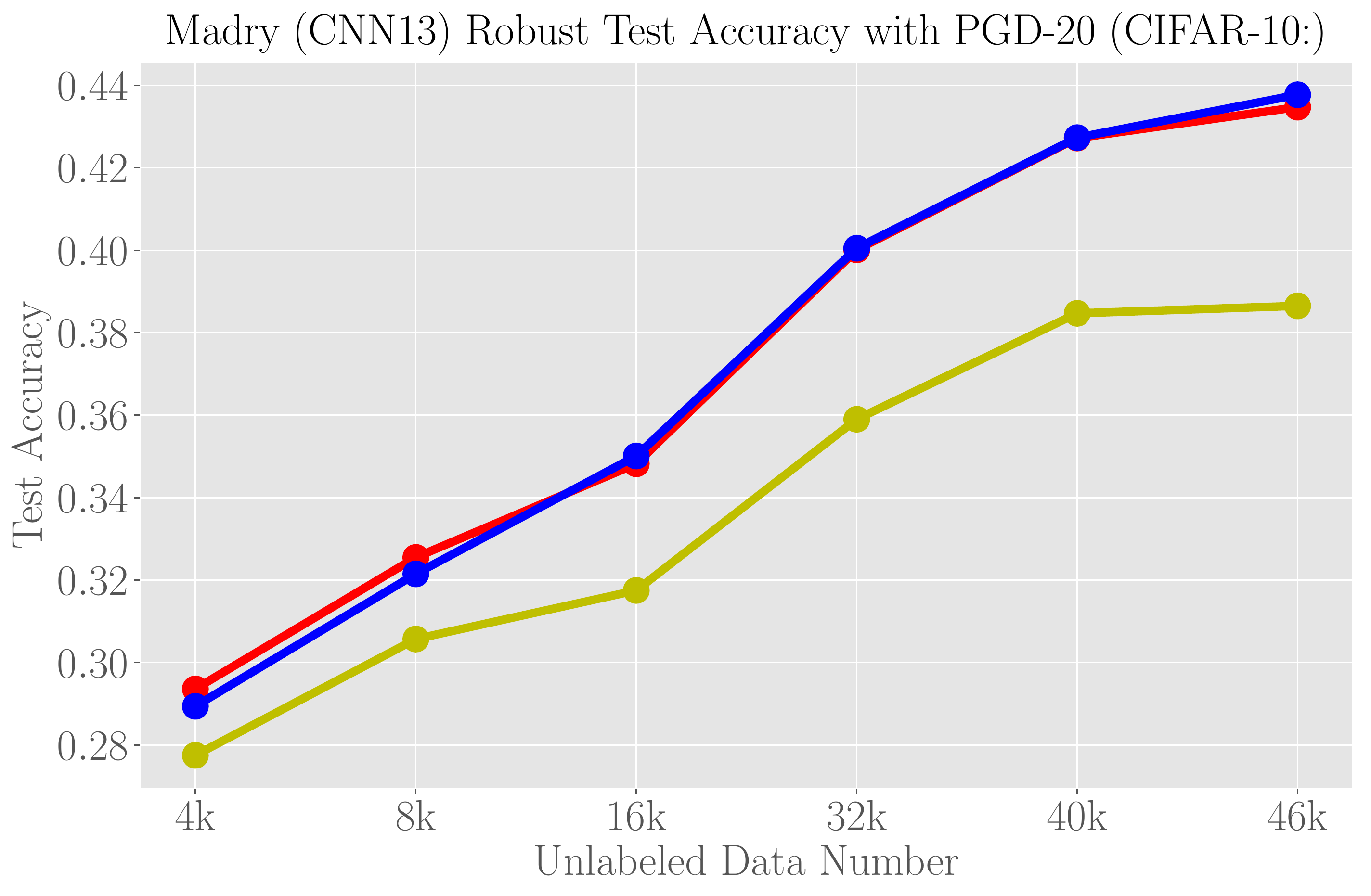} 
    \includegraphics[scale=0.12]{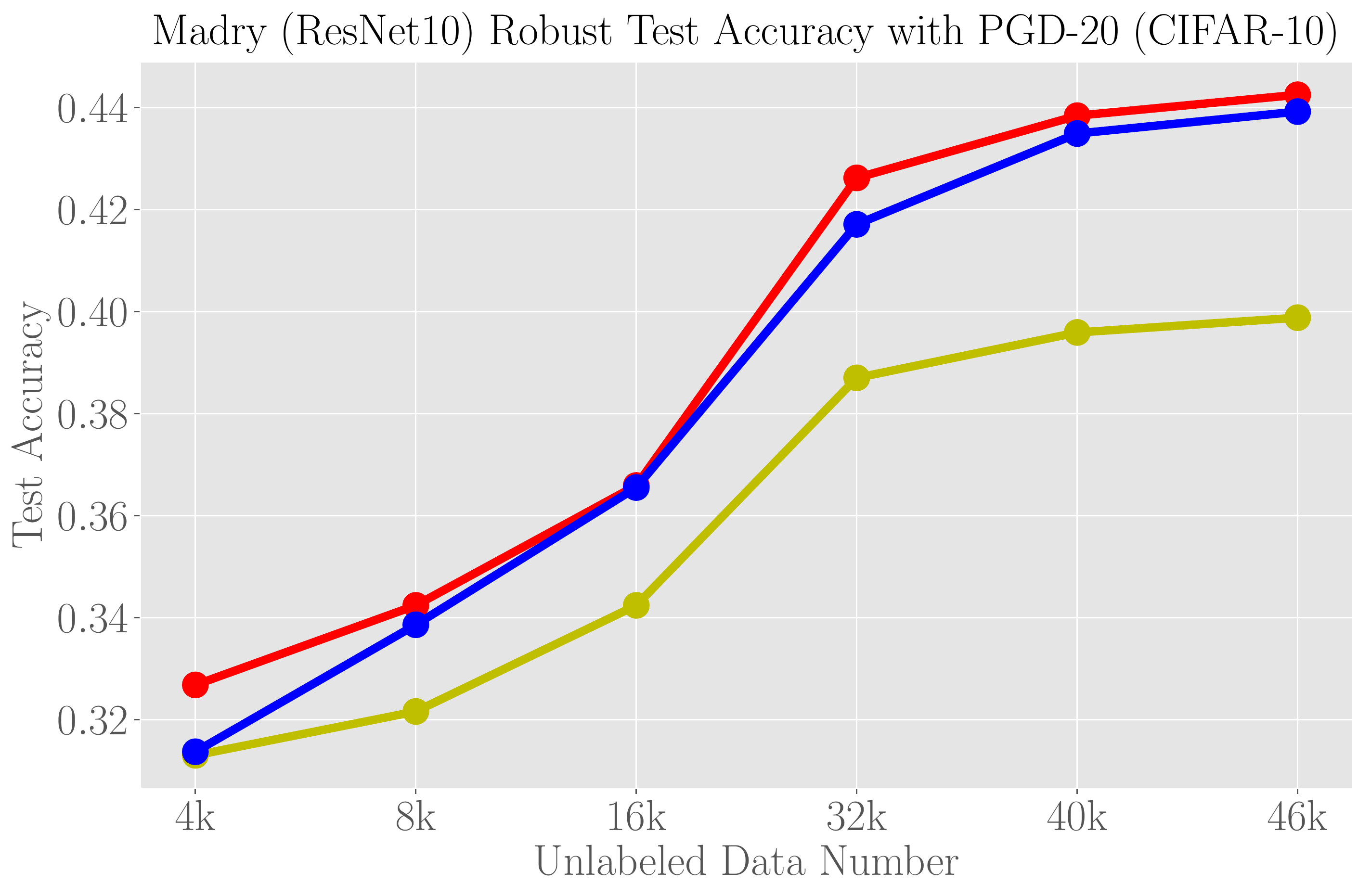} 
	\caption{
	%Performance of adversarial training based on datasets \textit{CIFAR-10} and \textit{SVHN}, two types of DNNs (CNN13 and ResNet10) and two types of adversarial training (TRADES and Madry).
	Performance comparisons between supervised oracle, robust self-training and our robust co-training. 
	The first column represents adversarial training TRADES on CNN13 on \textit{SVHN} dataset. The second column represents adversarial training TRADES on CNN13 on \textit{CIFAR-10} dataset. The third column represents Madry's adversarial training on CNN13 on \textit{CIFAR-10} dataset. The fourth column represents Madry's adversarial training on ResNet10 on \textit{CIFAR-10} dataset. The first row represents standard test accuracy evaluated by natural test data. The second row represents robust test accuracy evaluated by adversarial data (PGD-5). 
	The third row represents the robust test accuracy evaluated by adversarial data (PGD-20).} 
	\label{cifar_svhn_adv_training}
\end{figure}

We annotate unlabeled data $D_U$ to achieve $S_U = \{(\mathbf{x}_j, \Bar{y}_j)\}_{j=0}^{N_U - 1}$, where $\Bar{y}_j$ is pseudo label. Note that different methods can acquire $S_U$, such as pre-determined annotator (in RST~\cite{Percy}), deep co-training (in Algorithm~\ref{robust_co_adversarial_training}), and experts labeling.
%Adversarial training is conducted based on $S= S_L \cup S_U$.
Then, we combine 4000 labeled dataset $S_L$ with pseudo-labeled dataset $S_U$ into $S = S_L \cup S_U$, and conduct adversarial training based on $S$. In Figure~\ref{cifar_svhn_adv_training}, we use adversarial training TRADES~\cite{Zhang_trades} (i.e., Eq.~\eqref{trades}) or Madry's adversarial training~\cite{MadryMSTV18} (i.e., Eq.~\eqref{madry_adversarial_train}) to conduct experiments on CNN13 and ResNet10, where $\lambda$ in Eq.~\eqref{trades} is set to 1 for all experiments by TRADES. Both Madry's adversarial training and TRADES use PGD-10 to exploit adversarial examples. For \textit{CIFAR-10}, $\epsilon=0.031$ and step size is 0.007. For \textit{SVHN}, $\epsilon = 0.0156$ and step size is 0.007. Inputs are normalized between 0 and 1. 
To sum up, we compare three adversarial training methods leveraging unlabeled data in Figure~\ref{cifar_svhn_adv_training}.
\begin{itemize}
  \item Supervised oracle (red line): $S_U$ is labeled by experts achieving 100\% correct labels to all unlabeled data.
  \item Robust self-training (yellow line): $S_U$ is labeled by pre-determined annotator, which provides around 73\% correct labels on unlabeled \textit{CIFAR-10} data and around 82\% correct labels on unlabeled \textit{SVHN} data (yellow line in Figure~\ref{label_quality_compara}).
  \item Robust co-training (blue line): $S_U$ is labeled by deep co-training. Depending on the amount of unlabeled data, deep co-training could give around 80\% - 90\% correct labels to unlabeled \textit{CIFAR-10} data and around 89\% - 92\% correct labels on unlabeled SVHN data (blue line in Figure~\ref{label_quality_compara}).
\end{itemize}

To evaluate the performance, we calculate the standard test accuracy using natural test data, and robust test accuracy using its corresponding adversarial test data. Adversarial test data are generated by PGD-5 and PGD-20 respectively, with the $\epsilon=0.031$ and step size is 0.003. Figure~\ref{cifar_svhn_adv_training} shows that, in terms of different datasets, adversarial training methods and network structures, the quality improvement of pseudo labels can obviously improve adversarial training, namely both standard test accuracy and robust test accuracy get improved significantly.

%standard test accuracy \footnote{+++ However, you mention Robust Accuracy in Figure 6 instead of standard test accuracy. I think all your terms are not uniform.}, our robust co-adversarial training (blue line) achieves very good results, which are close to the supervised oracle. 
%By contrast, the pre-determined annotator adopted by RST give comparable worse results directly due to its bad labelling accuracy.
%In Figure~\ref{trades_cnn13_PGD-5_test_acc} and Figure~\ref{trades_cnn13_PGD-10_test_acc},
%We use ResNet10 to carry on robust training,  
%PGD-0 mean clean data test accuracy.\\
%PGD-5 mean attacks using projected gradient descent with step numbers 5 and step size is 0.003.\\
%PGD-10 mean attacks using projected gradient descent with step numbers 10 and step size is 0.003.
%Figure~\ref{trades_cnn13_PGD-5_test_acc} and Figure~\ref{trades_cnn13_PGD-10_test_acc} show that improved label quality in part (b) could significantly boost performance of robust training in term of robust test accuracy under various attacks, e.g. PGD-5 and PGD-10 
%\footnote{+++ Why only PGD-5 and PGD-10? It is a bit weak, normally PGD-20 and PGD-40.}.
%By contrast, using prd-determined annotator shows the little improvements on robust performance.
%\subsection{Abalation Study}
%\footnote{+++ This part should do like this, fix (a) (c) and change (b); fix (a) (b) and change (c). This part should be discussed with Gang. Our claim is to emphasize (b) step is the key.}
%Better to control different combinations (e.g., algorithm level, dataset level) to justify our claim..

\section{Conclusion}
In this paper, we investigate the bottleneck of adversarial learning with unlabeled data, and find the affirmative answer ``the quality of pseudo labels on unlabeled data". To break this bottleneck, we leverage deep co-training to boost the quality of pseudo labels, and thus propose robust co-training (RCT) for adversarial learning with unlabeled data. We conduct sufficient experiments on \textit{CIFAR-10} and \textit{SVHN} datasets. Empirical results demonstrate that RCT can significantly outperform robust self-training (RST) in both standard test accuracy and robust test accuracy w.r.t. different datasets, different network structures, and different adversarial training. In future, we will investigate theory of RCT, and explore more robust adversarial learning methods.

\subsection*{Acknowledgments}
MS was supported by JST CREST Grant Number JPMJCR1403.
\clearpage

\bibliographystyle{IEEEtran}
\bibliography{references}
\end{document}